\documentclass[]{fairmeta}
\usepackage{xcolor}         
\usepackage[table, dvipsnames]{xcolor}
\usepackage{wrapfig}
\usepackage{graphicx}
\usepackage{pifont}
\usepackage{multirow}
\usepackage{subcaption}

\usepackage{amsmath,amsfonts,bm}

\def\eqref#1{equation~\ref{#1}}

\def\1{\bm{1}}

\DeclareMathAlphabet{\mathsfit}{\encodingdefault}{\sfdefault}{m}{sl}
\SetMathAlphabet{\mathsfit}{bold}{\encodingdefault}{\sfdefault}{bx}{n}

\newcommand{\E}{\mathbb{E}}

\newcommand{\KL}{D_{\mathrm{KL}}}

\title{Information-Regularized Attention for Visual-Centric Reasoning}

\author[1,2,*]{Guohao Sun}
\author[1]{Xiaofang Wang}
\author[1]{Yash Patel}
\author[1]{Mengchen Liu}
\author[2]{Zhiqiang Tao}
\author[1]{Praveen Krishnan}
\affiliation[1]{FAIR at Meta}
\affiliation[2]{Rochester Institute of Technology}

\contribution[*]{Work done at Meta}

\abstract{Vision–language models (VLMs) have become a paradigm for multimodal learning, yet remain unstable due to object hallucination, weak visual grounding, and catastrophic forgetting after full-parameter instruction tuning. We claim these failures result from a lack of explicit control over visual representation learning during the standard next-token prediction objective. As a result, visual embeddings thus become passively optimized and prone to injecting redundant or spurious signals. To counter this, we introduce Information-Regularized Attention (IRA), a stochastic attention mechanism that explicitly regulates the amount of visual information injected into the hidden states of intermediate transformer layers. This local reparameterization translates uncertainty about visual representations into local noise that is independent across data points.
Beyond evaluating model performance, we also quantify embedding properties, where IRA produces smoother curvature trajectories and suppresses attention-sink across all layers, indicating a more stable transformation of the visual signal.
Our results suggest that stochastic attention is not merely a regularizer but a key contributor to representation learning in a generative architecture, offering a new direction for building more reliable VLMs.}

\date{\today}
\correspondence{Praveen Krishnan at \email{pkrishnan@meta.com}}

\begin{document}

\maketitle

\section{Introduction}
\label{sec:intro}
Vision-language models (VLMs) have emerged as a general-purpose framework for multimodal understanding, achieving strong performance across tasks such as visual question answering, image captioning, and multimodal dialogue~\cite{lu2019vilbert,alayrac2022flamingo,li2023blip,gan2022vision}. Despite this progress, modern VLMs remain limited by reliability issues, including object hallucination and unreliable grounding, where generated content is not supported by the visual input~\cite{li-etal-2023-evaluating,rohrbach-etal-2018-object}. These failures suggest that standard next-text-token prediction objectives do not sufficiently align vision and language information in latent space.

Current VLM training paradigms are largely data-centric, relying on increasingly diverse forms of post-training supervision, including visual instruction tuning~\cite{liu2023visual,sun2024sq}, preference optimization~\cite{ouyang2022training,rafailov2023direct,Peng_2025_ICCV,sun-etal-2024-self,Sun_2025_ICCV}, and policy optimization~\cite{Shao2024DeepSeekMathPT,schulman2017proximal,NEURIPS2025_95c6ae3f}. While effective, these approaches primarily improve model behavior by expanding the supervision signals, rather than directly regularizing the feature representations. Under the standard next-token prediction objective, all the embeddings are optimized only indirectly through language supervision. As a result, task-irrelevant or noisy visual signals can propagate through attention layers and interfere with cross-modal reasoning.

\begin{figure}[tb]
  \centering
    \includegraphics[width=0.85\linewidth]{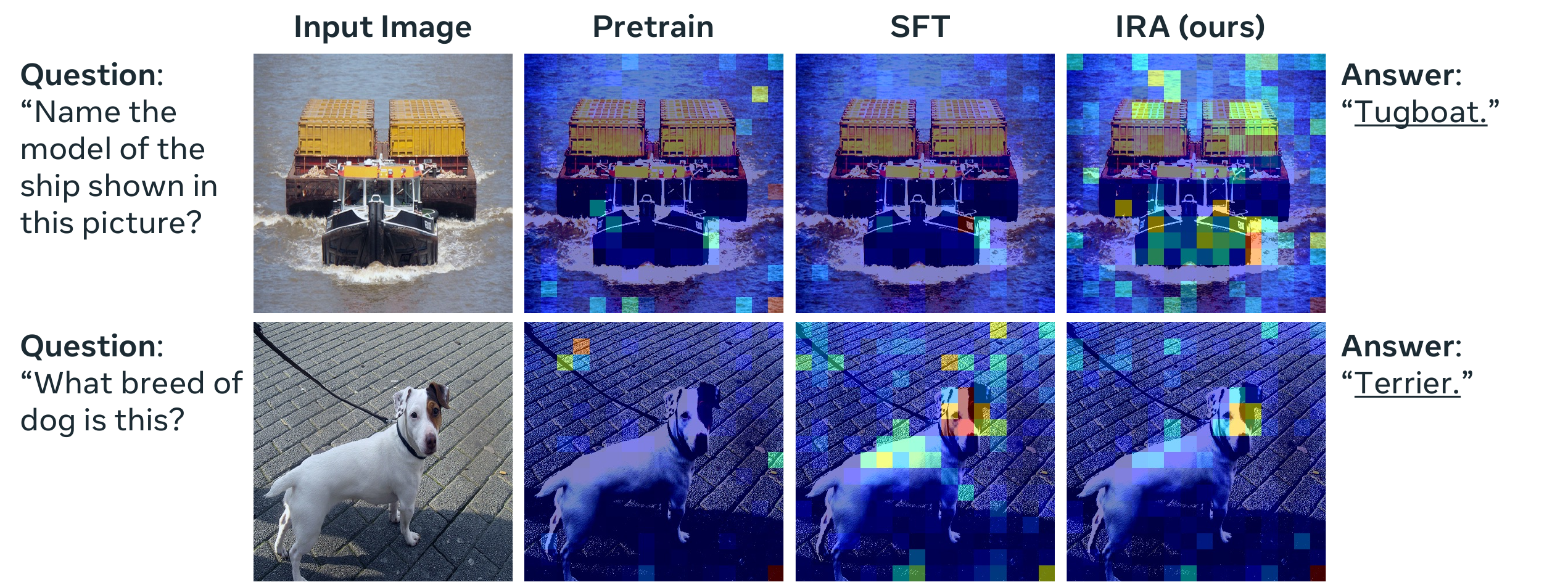}
  \caption{\label{fig:cover} Cross-attention of a VLM. We look at the normalized attention, i.e., $Attn(\text{answer} \rightarrow \text{visual tokens})$ of the last layer. The model indicates a bad visual dependency without visual instruction tuning. \textbf{SFT} improves alignment between visual cues and output text, but biased information is introduced due to unregularized visual embedding. \textbf{IRA} restricts biased knowledge and only encodes necessary information. }
\end{figure}

This issue is reflected in recent observations of attention sinks~\cite{gu2024attention,kang2025see,queipo-de-llano2026attention,barbero2025llms} and spike values in attention heads~\cite{sun2024massive,sun2026spike,xiao2023smoothquant}, where attention collapses onto semantically uninformative tokens and produces noisy cross-modal interactions~\cite{rohrbach-etal-2018-object,mahajan2025attention,jiang2025devils}. Our study in Fig.~\ref{fig:cover} provides further evidence: the pretrained VLM fails to attend to the relevant visual regions, whereas standard supervised instructional fine-tuning improves alignment only partially and still exhibits biased, noisy attention. These findings suggest that improving VLM reliability requires moving beyond data-centric post-training alone. Instead, we aim to improve visual representation learning within the generative architecture through end-to-end training, thereby achieving robust visual embeddings.

Prior work, including attention weight optimization~\cite{li2026reinforced,fan2020bayesian} and gating mechanisms~\cite{qiu2025gated}, has shown that controlling the information flow within the attention module can effectively reduce the attention sink. Despite their effectiveness, they operate at the level of attention weights or attention outputs rather than improving representation learning in the intermediate layers~\cite{belrose2023eliciting,jiang2025devils}. 
In contrast, we address the attention problem by introducing an information-theoretic mechanism that regularizes the attention input. We posit that hallucinations arise from a deficiency in visual mutual information~\cite{witten2020mini}: the token prediction is overwhelmed by \emph{noisy} visual tokens~\cite{kang2025see,Sun_2026_CVPR} that dilute the signal of query-relevant evidence. This occurs when the model fails to differentiate between semantically coherent and incoherent visual-textual pairs within its internal hidden representations~\cite{mahajan2025attention, skean2025layer}. 

To this end, we propose Information-Regularized Attention (IRA), a mechanism that explicitly regulates the amount of visual information injected into the visual hidden states within the attention module (see Fig.~\ref{fig:ira}). After evaluating embedding quality using geometric~\cite{hosseini2023large} measures (e.g., curvature), we find that representations that exhibit highly curved trajectories across transformer layers tend to produce unstable attention patterns and higher sink ratios, whereas smoother, more linear trajectories are associated with improved grounding and prediction. This suggests that embedding linearity is a key structural property underlying reliable representation learning. 

Our contributions are summarized as follows:
\begin{itemize}
    \item We propose a novel Information-Regularized Attention (IRA) mechanism that regulates visual representations before they enter cross-modal attention, explicitly controlling visual information at the representation level during full-parameter instruction tuning.
    \item We demonstrate that IRA improves the representation geometry, leading to greater robustness across tasks and stable training dynamics.
    \item We identify a correlation between the attention sink phenomenon and representational curvature, suggesting an avenue for future work to use representational geometry to design more robust multimodal architectures. 
\end{itemize}

\section{Related Works}
\subsection{Representation learning and neural dynamics.}
The evolution of general-purpose VLMs~\cite{li2024llava,liu2023visual,bai2023qwen,internvl} has focused on bridging vision encoders with LLMs. However, understanding how these models encode and organize information requires deeper analysis.
Early research utilized linear probes~\cite{alain2016understanding} and SVCCA~\cite{raghu2017svcca} to examine feature dynamics, though these were largely confined to vision-only or shallow networks. Recent literature has extended to layer-wise analysis in large-scale LLMs, revealing that linguistic features and semantic roles are most effectively encoded in the intermediate Transformer layers~\cite{skean2025layer}. Specifically, the middle layers contain surprisingly robust features~\cite{voita2019bottom}, challenging the traditional emphasis on final-layer representations.
Our work builds on this by regularizing the representations within intermediate layers to achieve robust embeddings.\\
\subsection{Information regularization.}
Regularization is a common technique to improve the generalization performance of deep neural networks, and various implementations are available depending on the network architecture and target application. A well-known example is dropout~\cite{10.5555/2627435.2670313}, which randomly turns off a subset of hidden units in neural networks by multiplying a Bernoulli-distributed noise term. On the other hand, information bottleneck~\cite{tishby2000information,alemi2017deep} formalizes representation learning as a principled trade-off between task sufficiency and information compression~\cite{queipo-de-llano2026attention,hong2025comprehensive}. Recent work has begun applying these principles to VLM robustness. For example, VIB-Probe~\cite{zhang2026vib} utilizes VIB to detect and mitigate hallucinations by filtering noise from internal attention heads. However, most of the works treat the latent variable as external knowledge to support the downstream task. Instead, this work uses the information bottleneck as a regularization tool, aiming to directly influence representation learning within the transformer layers from a generative model. \\
\subsection{Mitigating attention sinks.}
Recent studies have highlighted phenomena in transformer architectures that hinder reliability, namely attention sinks~\cite{gu2024attention, xiao2023efficient} and massive activations~\cite{sun2024massive}. In causal generative models (e.g., LLM and VLM), attention sink~\cite{kang2025see} occurs when the model attends to irrelevant tokens, often driven by extreme outliers in the hidden state. Current mitigation strategies include gating mechanisms~\cite{qiu2025gated} and attention distribution optimization~\cite{li2026reinforced, jiang2025devils,zhao-etal-2025-mitigating}. While these methods optimize the attention weights themselves, they often treat the underlying representations untouched. Our proposed IRA goes beyond attention distribution optimization by regularizing the inner representations prior to attention computation.

\section{Method}
\subsection{Information-Theoretic View of VLM}
A vision-language model (VLM) consists of a vision encoder, a projector, and a large language model (LLM). The input to the LLM is a concatenated sequence of visual $x_{img}$ and language tokens $x_{txt}$. We use $x$ to denote the set of all $img$ and $txt$ tokens concatenated together. Within the LLM, each transformer layer $f(\cdot)$ applies a deterministic mapping to produce hidden states $h^{(\ell+1)} = f(h^{(\ell)})$, where $h^{(0)} = x$. The final hidden states $h^{(L)}$ are decoded into output tokens $y$. Therefore, the entire model defines a Markov chain:
\begin{equation}\label{eq:chain}
x \rightarrow h^{(1)} \rightarrow \dots \rightarrow h^{(L)} \rightarrow y,
\end{equation}
and $p_\theta(y \mid h^{(L)})$ is the probability of final answer $y$ given output layer hidden states $h^{(L)}$, predicted by a model parameterized by $\theta$. 

After simplifying \eqref{eq:chain} into $x \rightarrow h \rightarrow y$, we can potentially formulate a mutual information between $h$, and $(x,y)$ as: 
\begin{align}\label{eq:mutual_all}
    \mathbb{I}(h;x,y) &=\mathbb{I}(h;x)+\mathbb{I}(h;y\mid x)\\
    &=\mathbb{I}(h;y)+\mathbb{I}(h;x\mid y)\,,
\end{align}
where $\mathbb{I}(h;x)=\mathbb{I}(h;y)+\mathbb{I}(h;x\mid y)-\mathbb{I}(h;y\mid x)$, and the IB objective becomes:
\begin{equation}\label{eq:ib_objective}
\max \mathbb{I}(h;y) - \beta \, \mathbb{I}(h;x\mid y)\,.
\end{equation}
Given this objective, we can learn a representation $h$ that preserves useful predictive information and suppresses nuisance input information.
However, $h$ is a deterministic mapping of $x$ in standard LLM, and the next-text-token prediction using supervised finetuning (SFT) optimizes  
\begin{equation}
    \max_\theta\E\left[\log p_\theta(y\mid h)\right]\approx \max_\theta \mathbb{I}(h;y) \,,
\end{equation}
without discarding irrelevant information in the condition feature $h$.
We hypothesize that using SFT alone may encode noisy information into visual embeddings, diluting attention and reducing discrimination among visual tokens. 

To address this challenge, we propose that information regularization can be formulated as injecting random noise into $h$, yielding stochastic hidden units $z$. By doing this, we can leverage well-defined probabilistic formulations to analyze the conventional training procedure and propose better optimization approaches.
\subsection{Information Regularization}
In this work, we formulate information regularization as a variational inference problem~\cite{noh2017regularizing} in latent space. 
Given the representations $h^{(\ell)}$ of input from each layer, we aim to sample a latent variable $z^{(\ell)}$ from a posterior distribution as:
\begin{equation}
    z^{(\ell)}\sim p(z^{(\ell)}\mid h^{(\ell)})\,,
\end{equation}
but approximating posterior is intractable through Bayes' rule $p(z\mid h)=\frac{p(h\mid z)p(z)}{p(h)}$. This work involves variational inference to approximate the intractable posterior. In this formulation, inference is cast as an optimization problem in which we optimize the model parameters $\phi$ of $q_\phi(z^{(\ell)} \mid h^{(\ell)})$ to approximate $p(z^{(\ell)}\mid h^{(\ell)})$. 

Such a variational principle provides a tractable lower bound as 
\begin{equation}
    \text{\L}(\theta,\phi|y)=\E\left[\log p_\theta(y\mid z^{(L)})\right]- \sum_{\ell=1}^{\mathcal{L}}\KL(q_\phi(z^{(\ell)}\mid h^{(\ell)})\,\|\,p(z^{(\ell)}))\,.
\end{equation}
This work reinterprets each layer as performing a data-dependent amortized variational inference. Specifically, instead of introducing a separate inference network $q_\phi(\cdot)$, we attach a lightweight parametric head to each transformer layer to produce the posterior as $q_{\theta,\phi}(z^{(\ell)}\mid h^{(\ell)})$, thereby tightly coupling latent variable encoding with the forward pass of the model. This design enables efficient layer-wise stochasticity without additional encoding overhead, and allows the variational parameters (i.e., $\phi$) to co-evolve with the backbone representations (i.e., $\theta$) during end-to-end training. 

\begin{figure}[t]
  \centering
    \includegraphics[width=0.35\linewidth]{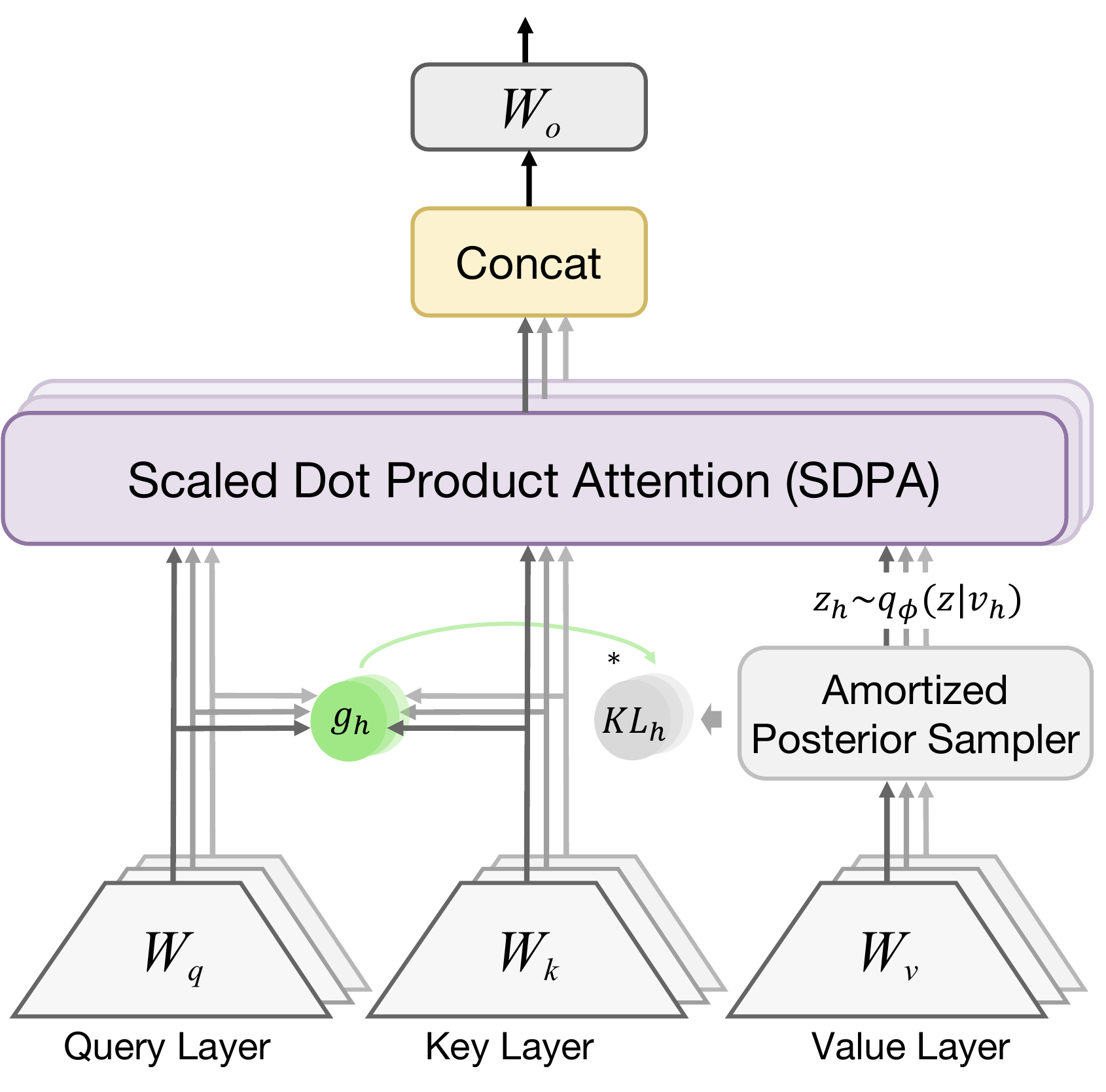}
  \caption{\label{fig:ira}Architecture of the proposed Information-Regularized Attention (IRA). We introduce a lightweight posterior sampler that incorporates stochastic representations prior to the attention computation. }
\end{figure}
\subsection{Regularizing Visual Representation in Attention Module}
Multi-head attention (MHA) plays a central role in embedding updates by enabling the model to process information through multiple parallel pathways~\cite{NIPS2017_3f5ee243}. Each attention head applies independent projections of \textbf{queries}, \textbf{keys}, and \textbf{values}, allowing the model to attend to different representation subspaces and relational structures simultaneously.
From an information-theoretic perspective, the value states $v$ are the primary channel through which information enters the output representation, where $v = hW_v$ with $v \in \mathbb{R}^{\mathcal{H}*d}$, $\mathcal{H}$ is the number of attention heads, and $d$ is the dimension for each. Therefore, instead of regularizing each transformer layer's output representation, we apply it before value states enter the attention computation, as shown in Fig.~\ref{fig:ira}. This mechanism enables the subsequent MLP layer to adapt smoothly to the stochastic attention output. This work mainly focuses on regularizing the visual representation by first extracting the visual components of the input value states.

Conceptually, the causal LLM in our VLM system can be viewed as a flexible architecture that shares parameters between the generative and inference processes, using the same hidden representations to produce outputs and to parameterize the variational posterior. This parameter sharing is reminiscent of hierarchical variational models such as Ladder-VAE~\cite{sonderby2016ladder}, although our formulation does not explicitly implement separate bottom-up and top-down inference pathways. Therefore, to alleviate the same early collapse and local-optimal issues~\cite {sonderby2016ladder}, we introduce a data-dependent prior and a prior-dependent posterior.
\subsubsection{Prior.}
This work defines the prior distribution of $z$ as data-dependent~\cite{Li2020Data-dependent,dziugaite2018data}, enabling adaptive regularization of latent variables conditioned on intermediate representations. However, such a design makes KL collapse to zero easily (i.e., $\KL(q_\theta(z\mid v)\,\|\,p(z\mid v))\approx0 $), since the posterior and prior both condition on the same input. To address this problem, we propose $p(z\mid \left \lfloor v\right \rfloor)$, where the input $v$ is detached from the gradient graph. This stop-gradient design prevents trivial posterior–prior collapse and ensures that the KL term acts as a stable regularization signal. 
Overall, the prior distribution of $z$ is a perturbation of $v$ parameterized by a Gaussian:
\begin{equation}
p_{\lambda}(z\mid \left \lfloor v\right \rfloor) = \mathcal{N}(  \left \lfloor v\right \rfloor , \sigma_p^2 I),
\end{equation}
where $\sigma_p^2=\exp(\log\sigma_\lambda^2)\in \mathbb{R}^{\mathcal{H}*d}$, and $\log\sigma_\lambda^2$ is a learnable parameter shared by all the latent variables from the same layer. This formulation can be interpreted as learning a zero-mean Gaussian noise from the entire training data and regularizing the intermediate representations of each data point, thereby enabling adaptive regularization while preserving the underlying semantic content.
\subsubsection{Posterior.}
The posterior is parameterized as an Isotropic Gaussian as:
\begin{equation}
q_{\theta,\phi}(z|v)=\mathcal{N}(v + \Delta_\phi(v), \sigma_q^2 I),
\end{equation}
where $\sigma_q^2 = \exp(\log\sigma^2_\phi(v))$, $\Delta_\phi(\cdot)$ and $\log\sigma^2_\phi(\cdot)$ are predicted by a linearly layer parameterized by $\phi$ condition on the current value state.
The mean residual $\Delta_\phi(v)\in \mathbb{R}^{\mathcal{H}*d}$ captures an adaptive perturbation to the pretrained representation $v$, allowing the model to refine visual features when beneficial for prediction. Meanwhile, $\sigma_q^2\in \mathbb{R}^{\mathcal{H}*1}$ controls the uncertainty of this refinement, determining how much flexibility is permitted for each attention head.
\subsubsection{KL regularization.}
The independent KL divergence between posterior and prior in each transformer layer becomes:
\begin{equation}\label{eq:klloss}
\KL(q_{\theta,\phi}\,\|\,p_{\lambda})
=\frac{1}{2}\left[\frac{\|\Delta_\phi(v)\|^2 
+ \sigma_q^2}{\sigma_p^2}- 1+ \log \sigma_p^2- \log \sigma_q^2\right].
\end{equation}
The KL regularization constrains the posterior to remain close to a detached, reference distribution centered at the original value state. The mean-shift penalty limits excessive deviation from the pretrained representation, while the variance-matching terms regulate the magnitude of injected stochasticity. 
To propagate gradients back through $q_\phi$, we use the reparameterization trick as:
\begin{equation}\label{eq:reparam}
z=v + \Delta_\phi(v)+\sigma_q*\epsilon, \quad\epsilon \sim \mathcal{N}(0, I)\,.
\end{equation}
This denotes as injecting learnable feature-level Gaussian noise (i.e., $\Delta_\phi(v)+\sigma_q*\epsilon$) in the deterministic value states $v$.



\subsection{Uncertainty-Conditioned Information Regularization}
The KL regularization in \eqref{eq:klloss} applies a uniform penalty across all visual tokens and attention heads, regardless of their semantic importance. However, in multi-head attention, different heads specialize in different types of visual relationships. Some heads may focus on spatial alignment, others on semantic attributes, and others on fine-grained textures.
This observation suggests that regularization should be applied selectively rather than uniformly. Specifically, more should be applied to visual tokens that are both: i) \textbf{Unimportant}, where attention assigns low relevance, and ii) \textbf{Uncertain}, where cross-modal attention distribution has high entropy. We therefore introduce a weighting mechanism that adaptively adjusts the regularization magnitude for each visual token.
Let $q_{txt}$ and $k_{img}$ denote the text and visual parts of the query and key states, respectively. We then compute the normalized attention distribution as:
\begin{equation}
p = \operatorname{softmax}(\frac{\left \lfloor q_{txt} \right \rfloor^{\top} \left \lfloor k_{img} \right \rfloor}{\sqrt{d}})\,,
\end{equation}
where $ \left \lfloor \cdot \right \rfloor$ is gradient detachment, preventing the KL regularization term from directly influencing $q$ and $k$ representations, which could otherwise lead to unstable optimization. 

Given $p\in\mathbb{R}^{\mathcal{T}*\mathcal{S}*\mathcal{H}}$, where $\mathcal{T}$ and $\mathcal{S}$ are the number of text and visual tokens, we compute the un-normalized importance score of the $i$-th visual token as:
\begin{equation}
a_{i}=\frac{1}{\mathcal{T}}\sum_{j=1}^\mathcal{T} p_{j,i}\,,
\end{equation}
where $a_{i}\in\mathbb{R}^{\mathcal{H}}$ indicates the average attention mass assigned to the $i$-th visual token by the language contexts in each head. However, relying solely on high or low importance scores is unreliable, as the softmax function's normalization often leads to an attention sink~\cite{barbero2025llms,gu2024attention}. In particular, high attention weights do not necessarily indicate semantically meaningful information, as they may arise from biased distributions. To address this limitation, we propose reweighting token importance based on visual attention entropy. Entropy quantifies the dispersion of attention and provides a signal indicating the confidence of attention allocation. Intuitively, a low-entropy distribution reflects confidence and requires less regularization, whereas a high-entropy distribution suggests a potentially noisy representation and thus requires more regularization.
To achieve this goal, we compute normalized entropy as:
\begin{equation}\label{eq:uncertainty}
\mathbb{H}=\frac{1}{\mathcal{H}}\sum_{n=1}^{\mathcal{H}}-\frac{\sum_{j=1}^\mathcal{T} p_{j,n} \log p_{j,n}}{\log \mathcal{T}}\,.
\end{equation}
To this end, we construct a token-wise weighting mechanism (i.e., $g_{i} = \mathbb{H} \cdot (1 - a_{i})$) to adaptively weight the KL penalty defined in \eqref{eq:klloss} for the $i$-th stochastic visual token as:
\begin{equation*}
    \KL(q_{\theta,\phi}(z\mid v)\,\|\,p_{\lambda}(z\mid \left \lfloor v\right \rfloor))=\E_{(i)\sim Unif}\left[g_{i}*\KL(q_{\theta,\phi}(z_{i}\mid v_{i})\,\|\,p_{\lambda}(z_{i}\mid \left \lfloor v_i\right \rfloor))\right]\,.
\end{equation*}

Moreover, we control the magnitude of noise by $z_{i}=v_{i} + \Delta_\phi(v_{i})+g_{i}*\sigma_{i}*\epsilon, \quad\epsilon \sim \mathcal{N}(0, I)$, preventing excessive noise injection into relatively important visual representations and encourage stronger regularization on the others. Notably, the weighting estimate $g$ is only required during training, and we use $z=v + \Delta_\phi(v)$ at inference time.

Overall, the final objective function becomes: 
\begin{equation*}
    \text{\L}(\theta,\phi,\lambda|y)=\E\left[\log p_\theta(y\mid z^{(\mathcal{L})})\right]- \beta\sum_{\ell=1}^{\mathcal{L}}\KL(q_{\theta,\phi}(z^{(\ell)}\mid v^{(\ell)})\,\|\,p_\lambda(z^{(\ell)}\mid \left \lfloor v^{(\ell)}\right \rfloor))\,,
\end{equation*}
where $\beta$ is scheduled using cosine interpolation, increasing from 0 to $\beta_{max}$ during the first $k\%$ of the total training steps. We conduct experiments of $\beta_{max}$ for a better regularization-performance trade-off in Appendix~\ref{appen:ana2}. Typically, we find $\beta_{max}=1\times 10^{-4}$ and $k=50$ achieves the best performance.
By initializing training with only reconstruction loss and gradually introducing the variational regularization term, we avoid local minima at $\KL(q\,\|\,p)\approx 0$ during optimization. This idea has previously been considered in~\cite{sonderby2016ladder,bowman2016generating}.  

\begin{table*}[h]\setlength{\tabcolsep}{2pt}
\centering
\caption{\label{tab:hyperp} Hyper-parameters for training the models in different stages. We denote Stage 1/1.5 as pre-training (\textbf{Pt}) and Stage 2 as full-parameter supervised fine-tuning (\textbf{SFT}). The proposed \textbf{IRA} is an improvement of \textbf{SFT} in \textbf{Stage 2} training.}
\scalebox{0.7}{
\begin{NiceTabular}{ll|ccccccccc}
\toprule
 & Model & Trainable & \#Data & Batch size & Context len & LR (Backbone) & LR (IRA) & W Decay & Epochs & IRA \\ \midrule
\multirow{2}{*}{Stage 1} & InternVL2.5-8B & \multirow{2}{*}{MLP} & - & 512 & 16384 & $2\times 10^{-4}$ & - & 0.05 & - & \multirow{2}{*}{No} \\
 & LLaVA-OneVision-8B &  & 558k & 512 & 8182 & $2\times 10^{-6}$ / $1\times 10^{-5}$ & - & 0 & 1 &  \\ \midrule
\multirow{2}{*}{Stage 1.5} & InternVL2.5-8B & ViT + MLP & - & 512 & 16384 & $1\times 10^{-5}$ & - & 0.05 & - & \multirow{2}{*}{No} \\
 & LLaVA-OneVision-8B & Full Model & 4M & 256 & 32768 & $2\times 10^{-6}$/$1\times 10^{-5}$ & - & 0 & 1 &  \\\midrule
\multirow{3}{*}{\begin{tabular}[c]{@{}l@{}}\textbf{Stage 2}\\ (ours)\end{tabular}} & InternVL2-8B & \multirow{3}{*}{Full Model} & \multirow{3}{*}{3.2M} & 512 & 16384 & $1\times 10^{-5}$ & $1\times 10^{-4}$ & 0.05/0.01 & \multirow{3}{*}{1} & \multirow{3}{*}{Yes} \\
 & InternVL2.5-8B &  &  & 512 & 16384 & $1\times 10^{-5}$ & $1\times 10^{-4}$ & 0.05/0.01 &  &  \\
 & LLaVA-OneVision-8B &  &  & 256 & 32768 & $2\times 10^{-6}$/$1\times 10^{-5}$ & $1\times 10^{-4}$ & 0 &  &  \\ \bottomrule
\end{NiceTabular}}
\end{table*}
\section{Experiments}
\subsection{Experimental Setup}
\noindent\textbf{Model architectures.} We mainly conduct experiments on open-sourced VLMs, including InternVL2, InternVL2.5~\cite{internvl} and LLaVA-Onevision~\cite{li2024llava}. In practice, we chose these models because they encode visual input into a token sequence that can be mapped back to the pixel level, enabling us to study visual grounding with greater interpretability.\\
\noindent\textbf{Baseline method.} Previous works have empirically shown that supervised finetuning (SFT) has poor generalization and severe catastrophic forgetting~\cite{zhai2023investigating,wu-etal-2025-mitigating-visual,chen-etal-2020-recall} when models are adapted to diverse tasks and domains after full-parameter instruction tuning. Please see Appendix~\ref{appen:forget} for our empirical analysis of catastrophic forgetting. This tension raises a fundamental challenge for robust representation learning~\cite{dong2021should,poole2014analyzing,noh2017regularizing} and visual understanding. \\
\noindent\textbf{Training details.} Typically, training a VLM from scratch requires first \emph{pretraining} on image-text pairs and then full-parameter \emph{supervised fine-tuning} on visual instructional data, denoted as \textbf{Pt} and \textbf{SFT}. The IRA aims to improve full-parameter training during the SFT stage by initializing the model with the stage-1.5 checkpoint, as shown in Table~\ref{tab:hyperp}. Note that we set a 10$\times$ higher learning rate for the IRA parameters, allowing the regularization term to quickly catch up with task optimization.
At the model architecture level, IRA introduces negligible additional parameters to the attention module, including a \texttt{Linear}($d$, $d$+1) and an \texttt{Embedding}($\mathcal{H}$, $ d$) per layer, where $d=128$ and $\mathcal{H}=8$ indicate head dimension and $\#$heads. Such a modification transforms the deterministic propagation of visual embeddings into a stochastic process.\\
\noindent\textbf{Training data.}
We mainly follow LLaVA-Onevision~\cite{li2024llava} to prepare our training data, but we consider only the single-image subset with 3.2M samples, as the goal is to evaluate the generalization of SFT and IRA to multi-image and video understanding after training. We aim to test how different training methods mitigate forgetting and evaluate their robustness in an OOD scenario.

\begin{table}[t]
\centering
\caption{Comparison of training methods for general-purpose VLMs. We report the summation of perception and cognition scores for MME. The best results are \textbf{bold}.}\label{tab:generalQA}
\setlength{\tabcolsep}{2pt}
\scalebox{0.8}{
\begin{NiceTabular}{lcccccccccc}
\toprule
\multirow{2}{*}{Method} & \multicolumn{2}{c}{STEM} & \multicolumn{5}{c}{General QA} & \multicolumn{3}{c}{Text \& Chart} \\
\cmidrule(lr){2-3}\cmidrule(lr){4-8}\cmidrule(lr){9-11}
 & MMMU-Pro & MMMU & MME &MME-RW& MMStar & MMBench&OK-VQA & TextVQA& ChartQA&DocVQA\\
\midrule
InternVL2-Pt & 26.0 & 42.1 & 1598 &31.2& 43.7 & 70.9&\textbf{44.3}& 64.8&70.3&76.2 \\
+ SFT &30.1&45.1&1792&37.6&\textbf{60.2}&80.4&42.3&73.0&\textbf{80.1}&85.3\\
\rowcolor{metabg}+ IRA&\textbf{31.2}&\textbf{45.7}&\textbf{1959}&\textbf{40.5}&58.8&\textbf{81.0}&43.0&\textbf{73.8}&79.9&\textbf{85.5}  \\
\midrule
InternVL2.5-Pt & 26.7&41.3&1669&29.5&47.3&72.7&43.4&69.2&73.1&81.8\\
+ SFT &30.4&46.4&1981&40.4&\textbf{61.1}&80.6&39.8&74.5&81.0&\textbf{86.7}\\
\rowcolor{metabg}+ IRA&\textbf{30.6}&\textbf{47.6}&\textbf{2038}&\textbf{40.8}&58.8&\textbf{81.8}&\textbf{43.6}&\textbf{74.7}&\textbf{81.8}&\textbf{86.7}  \\
\midrule
LLaVA-OV-Pt &20.5&36.3&1530&28.0&30.6&67.3&11.0&36.8&37.6&11.5\\
+ SFT &27.4&44.4&2030&\textbf{40.9}&57.9&79.5&41.7&75.9&79.9&88.3 \\
\rowcolor{metabg}+ IRA &\textbf{28.0}&\textbf{45.3}&\textbf{2109}&40.0&\textbf{58.1}&\textbf{79.6}&\textbf{47.4}&\textbf{77.1}&\textbf{80.7}&\textbf{88.5} \\
\bottomrule
\end{NiceTabular}}
\end{table}

\subsection{General Visual Understanding}
VLM is a general-purpose assistant in the wild. To validate the capabilities in real-world scenarios with open-form instructions, we use MMStar~\cite{mmstar}, MME~\cite{Fu2023MMEAC}, MME-Realworld~\cite{zhang2024mme}, and RealWorldQA~\cite{realworldqa}. Beyond chat capability, visual perception assesses the model's reasoning ability, so we adopt MMMU~\cite{Yue2023MMMUAM} and MMMU-Pro~\cite{yue2025mmmu}. Previous studies have shown that generalization usually improves semantic reasoning but harms fine-grained visual understanding(e.g., text parsing)~\cite{jiang2024correlation,steinberg2026vision}. Therefore, we incorporate TextVQA~\cite{textvqa}, ChartQA~\cite{masry-etal-2022-chartqa}, DocVQA~\cite{mathew2021docvqa}, and EmbSpatial~\cite{du2024embspatial}. 

As shown in Table~\ref{tab:generalQA}, the improvements of IRA focus on reasoning-intensive benchmarks (STEM and General QA), indicating that IRA primarily enhances compositional reasoning and the robustness of cross-modal grounding rather than merely strengthening surface-level alignment. This suggests that the method is not acting as a generic regularizer but rather specifically improving how visual evidence is injected and used during reasoning. We observe that IRA yields the largest gains on knowledge-intensive benchmarks, OK-VQA~\cite{marino2019ok}, suggesting that noisy Attention improves semantic abstraction and cross-modal reasoning.
Text-intensive tasks require high-fidelity visual transmission, where nearly all signals are informative. The improvements on TextVQA, DocVQA, and ChartQA indicate that the proposed uncertainty-conditioned noise successfully prevents over-regularization of critical visual details.

\subsection{Robustness and Generalization}
We evaluate the model's robustness using four benchmarks. POPE~\cite{li2023evaluating} formulates hallucination detection as a binary object-existence verification task, enabling precise measurement of false positives induced by language priors. HallusionBench~\cite{guan2024hallusionbench} further probes fine-grained visual reasoning and illusion-induced errors through paired yes/no questions that require consistent, evidence-based judgments. VLM-are-biased~\cite{vo2025vision} and VLM-are-blind~\cite{rahmanzadehgervi2024vision} evaluate the model's ability to actually see low-level vision rather than relying on biased prior knowledge or reasoning. As shown in Table~\ref{tab:robustness}, IRA achieves improved robustness on the InternVL family model.

\begin{table}[t]
\centering
\caption{Comparison of robustness and generalization. }\label{tab:robustness}
\setlength{\tabcolsep}{4pt}
\scalebox{0.8}{
\begin{NiceTabular}{lcccccccc}\toprule
\multirow{2}{*}{Method} & \multicolumn{3}{c}{Multi-image and Video} & \multicolumn{4}{c}{Robustness} & Spatial \\
 \cmidrule(lr){2-4} \cmidrule(lr){5-8}\cmidrule(lr){9-9}
 & MuirBench & BLINK & MVBench & POPE & HallBench & VLM-Bias & VLM-Blind & EmbSpatial \\
\midrule
InternVL2-Pt & 31.1 & 41.1 & 35.6 & 83.9 & 33.6 & 17.6 & \textbf{39.0} & 52.3 \\
+ SFT & \textbf{38.3} & 45.1 & 48.7 & 86.4 & 35.7 & 18.1 & 33.4 & 63.9 \\
\rowcolor{metabg}+ IRA & 36.8 & \textbf{45.3} & \textbf{48.9} & \textbf{86.5} & \textbf{36.0} & \textbf{18.4} & 33.9 & \textbf{64.9} \\ \midrule
InternVL2.5-Pt & 31.6 & 39.5 & 47.4 & 86.6 & 32.5 & 18.3 & 36.4 & 51.7 \\
+ SFT & 35.2 & \textbf{47.1} & 51.5 & 87.0 & 37.6 & 17.8 & 33.9 & \textbf{64.5} \\
\rowcolor{metabg}+ IRA & \textbf{35.4} & 45.5 & \textbf{52.0} & \textbf{87.5} & \textbf{37.9} & \textbf{18.3} & \textbf{37.3} & 64.2 \\ \midrule
LLaVA-OV-Pt & 37.4 & 24.6 & 41.9 & 85.7 & 21.7 & 16.9 & 5.8 & 51.4 \\
+ SFT & 40.9 & 42.9 & 53.0 & 88.6 & \textbf{38.1} & \textbf{20.9} & 19.0 & \textbf{63.5} \\
\rowcolor{metabg}+ IRA & \textbf{41.4} & \textbf{43.2} & \textbf{54.0} & \textbf{88.7} & 37.0 & 19.7 & \textbf{19.1} & \textbf{63.5} \\
\bottomrule
\end{NiceTabular}}
\end{table}

To further assess the generalizability of each method, we evaluate its performance on OOD tasks. For multi-image understanding, we report MuirBench~\cite{Wang2024MuirBenchAC} and BLINK~\cite{fu2024blink}, and for video understanding, we use MVBench~\cite{li2024mvbench}. In Table~\ref{tab:robustness}, IRA consistently improves video understanding and competitive performance on the multi-image tasks. 
\begin{figure}[t]
  \centering 
  \begin{subfigure}{0.32\linewidth}
  \centering
    \includegraphics[width=\linewidth]{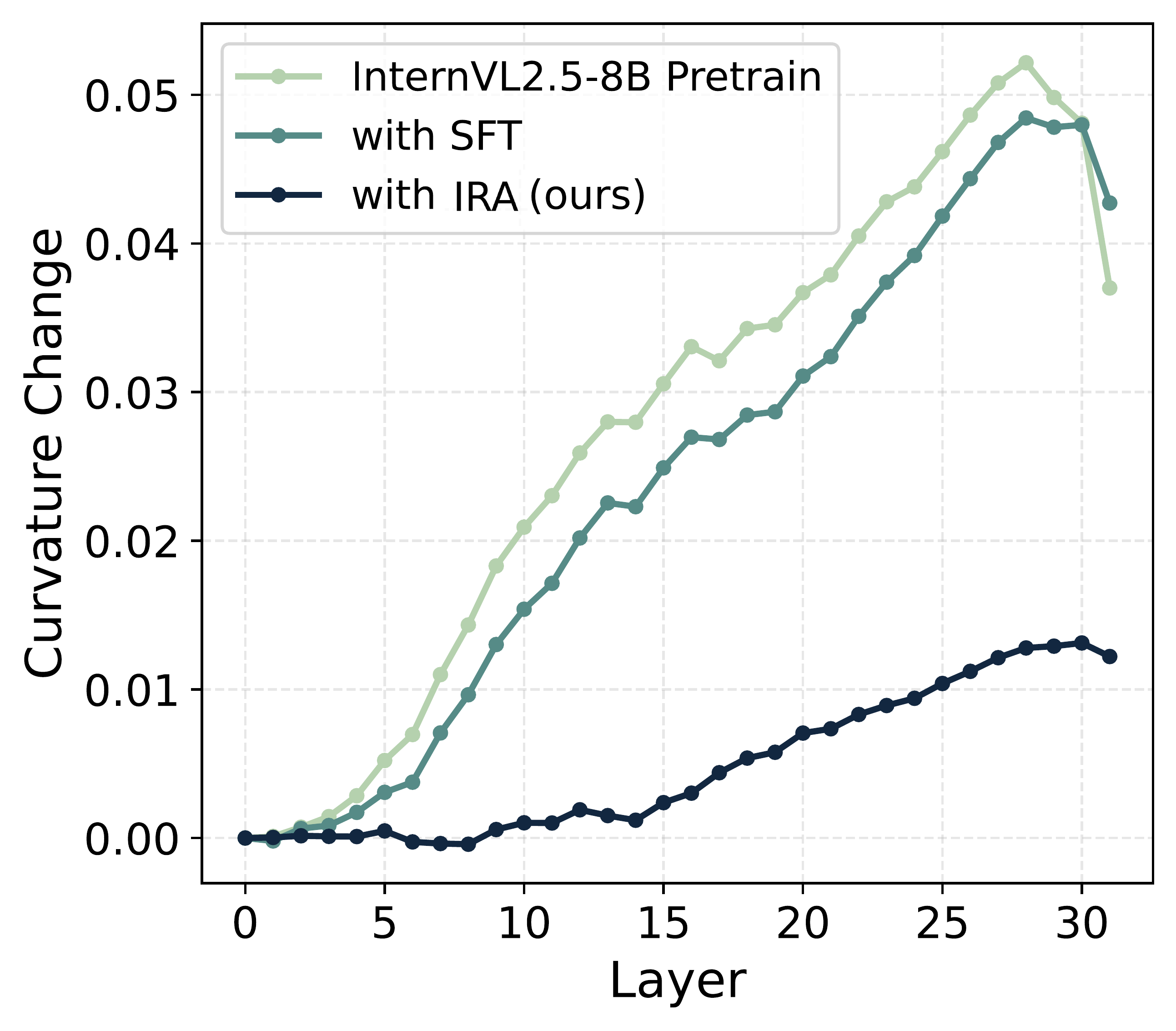}
    \caption{Curvature Change}
  \end{subfigure}
  \hfill
  \begin{subfigure}{0.32\linewidth}
  \centering
    \includegraphics[width=\linewidth]{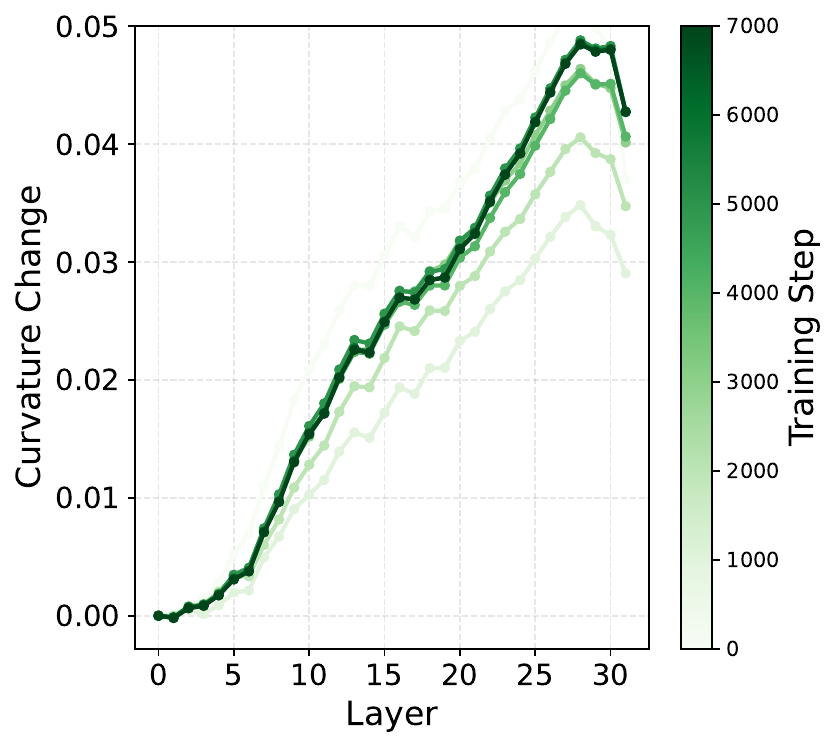}
    \caption{InternVL2.5-SFT}
  \end{subfigure}
  \hfill
  \begin{subfigure}{0.32\linewidth}
  \centering
    \includegraphics[width=\linewidth]{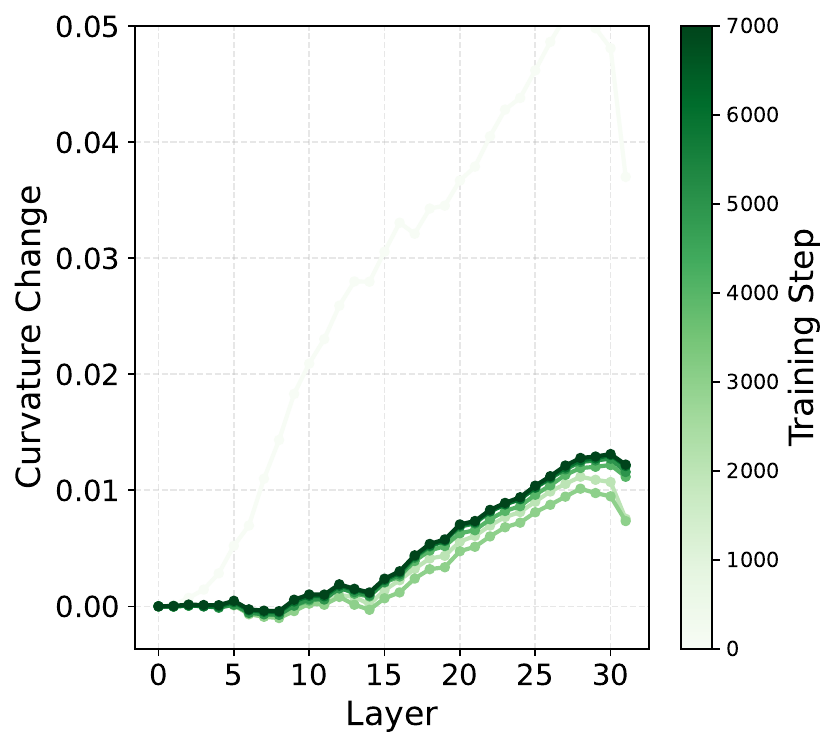}
    \caption{InternVL2.5-IRA}
  \end{subfigure}
  \caption{\label{fig:curvature}Analysis of the representation straightening. The y-axis shows the $\Delta C$ of each layer. A straighter trajectory indicates a smoother update to the representation. A model with better embedding quality exhibits a straighter curvature trajectory.  }
\end{figure}
\subsection{Representation Evaluation}
The brain transforms the incoming visual input to make it more predictable~\cite{henaff2019perceptual}. Predictive models operate by extrapolating future states from current representations, a process that is well-conditioned when internal representations evolve approximately linearly across model depth~\cite{hosseini2023large}. Inspired by these works, we developed a curvature metric based on neural trajectories of image tokens and used it in all analyses. Considering at layer $\ell$, we first extract the visual embeddings $x^{(\ell)}_1,x^{(\ell)}_2 \cdots x^{(\ell)}_I$ and compute $v^{(\ell)}_1,v^{(\ell)}_2 \cdots v^{(\ell)}_{I-1}$ as the difference between two adjacent states $v^{(\ell)}_k=x^{(\ell)}_{k+1}-x^{(\ell)}_k$. We then compute curvature as the angle between these vectors as
    ${\scriptstyle c^{(\ell)}_k=\text{arccos}(\frac{v^{(\ell)}_{k+1}\cdot v^{(\ell)}_k}{\left\| v^{(\ell)}_{k+1}\right\|\left\| v^{(\ell)}_{k}\right\|})}\,.$
Then the average curvature across the visual token sequence is computed as $C^{(\ell)}=\frac{1}{K}\sum_{i=1}^{K}c^{(\ell)}_i$. Finally, we compute a change in curvature between each layer and the first layer as $\Delta C^{(\ell)}=C^{(\ell)}-C^{(0)}$.
Our key insight is that curvature serves as a proxy for predictability: a straighter curvature trajectory across layers corresponds to smoother, more stable representational updates. As shown in Fig.~\ref{fig:curvature}, our method suppresses irregular variations and enforces straighter trajectories. Empirically, we show that reducing curvature is associated with improved training stability and attention distribution.

\subsection{Ablation Study}
The proposed IRA contains two key components: an uncertainty-guided weighting mechanism that adaptively controls the regularization, and a data-dependent prior centered on the pretrained visual representation. We conduct ablation studies in Table~\ref{tab:ablation} and observe that removing the weighting gate consistently degrades performance across most benchmarks, with particularly large drops on MMMU and MuirBench. This suggests that applying a uniform noise injection is too aggressive, while the weighting mechanism effectively balances the regularization of uncertain visual representations.
We study the effect of the prior design by replacing the data-dependent prior with a global learnable embedding initialized to zero. This modification consistently reduces performance, especially on text-intensive benchmarks, indicating that anchoring the prior to pretrained visual representations can save fine-grained visual information.

\begin{table*}[t]\setlength{\tabcolsep}{2pt}
\centering
\caption{\label{tab:ablation}Ablation study of our main designs. w/o IRA indicates vanilla SFT. }
\scalebox{0.8}{
\begin{NiceTabular}{lcccccccc}
\toprule
 Method& MMMU & RealWorldQA  & VQA$^{text}$&ChartQA & EmbSpatial & MuirBench & CVBench&Avg. \\ \midrule
\rowcolor{metabg}IRA&46.1&58.3&70.3&79.8&64.7&38.4&71.7&\textbf{61.3}\\
~~w/o weighted KL &45.1&57.0&69.5&80.4&65.6&35.9&71.8&60.7 \\
~~w/o $\mu_p=v$ &45.6&58.2&69.4&79.6&64.6&34.6&72.5&60.6 \\
~~w/o IRA&44.9&56.6&69.6&76.5&63.5&34.7&70.7&59.5 \\
\bottomrule
\end{NiceTabular}}
\end{table*}

\begin{figure}[t]
  \centering 
  \begin{subfigure}{0.42\linewidth}
  \centering
    \includegraphics[width=\linewidth]{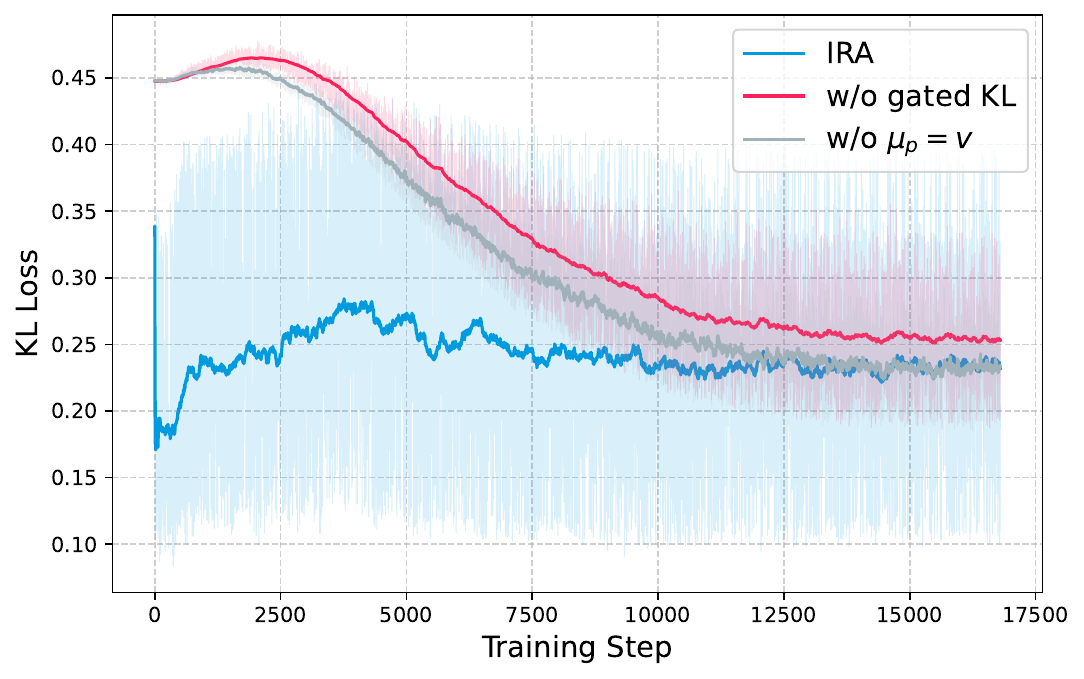}
    \caption{KL Loss During Training}\label{fig:klloss}
  \end{subfigure}
  \hfill
  \begin{subfigure}{0.57\linewidth}
  \centering
    \includegraphics[width=0.92\linewidth]{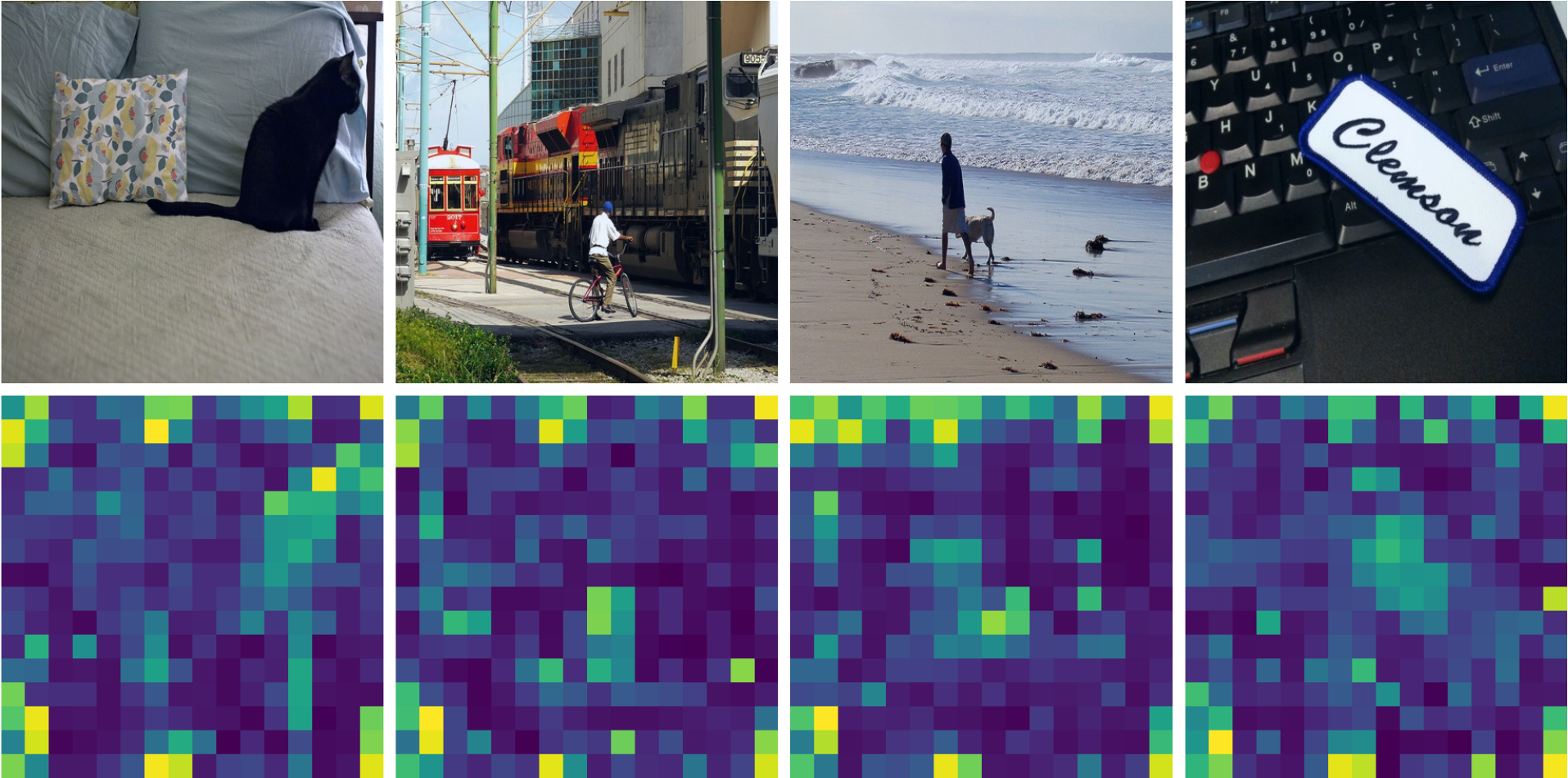}
    \caption{Visual Token Information Map}\label{fig:klDist}
  \end{subfigure}
  \caption{(a) Adaptive KL enables fewer regularization at the start of training, and gradually regularizes the prior knowledge. (b) Token-wise KL divergence across visual tokens, where brighter regions indicate stronger deviation from the prior and higher information capacity allocated to the corresponding visual patches. }
\end{figure}

Moreover, we analyze the KL loss during training in Fig.~\ref{fig:klloss}, where IRA maintains stable KL regularization throughout training, whereas removing the weighted KL results in higher KL values. This indicates that the weighting mechanism effectively prevents excessive regularization during the early stages of training. In contrast, replacing the pretrained-centered prior with a learnable prior also leads to greater KL fluctuations, suggesting that anchoring the prior to pretrained visual representations stabilizes the regularization process. After training, we visualize the KL divergence between the posterior distribution of each visual token and the global prior in Fig.~\ref{fig:klDist}, where brighter means higher KL, indicating more information. We have observed that the main objects in the image tend to be more important, while the background is less important. We find that some border tokens exhibit unexpectedly high importance, which may be due to their low frequency in the entire training data.

\begin{table*}[t]\setlength{\tabcolsep}{3pt}
\centering
\caption{\label{tab:layerstudy}Study of applying IRA in different layers. Assuming an LLM with 32 layers, $20\%-80\%$ means applying IRA in transformer layers 6-26. We utilize the $60\%-80\%$ setting in all experiments. }
\scalebox{0.8}{
\begin{NiceTabular}{lccccccccc}
\toprule
 Layer Depth&$\beta_{max}$& MMMU & MME  & VQA$^{text}$&ChartQA & EmbSpatial & MuirBench & CVBench&Improve($\uparrow\%$) \\ \midrule
 baseline&-&45.1&1839&72.0&78.4&63.6&37.2&72.1&+0.00\%\\ \midrule
$0\%-100\%$&$4\times10^{-4}$& 47.3&1940&70.0&76.0&61.2&34.4&72.3&-0.93\%\\
$20\%-80\%$&$4\times10^{-4}$&45.1&1956&70.0&75.7&61.0&35.0&68.7&-2.08\%  \\
$20\%-40\%$&$8\times10^{-5}$&43.4&1791&70.6&77.4&62.9&38.2&72.7&-1.30\%  \\
 $40\%-60\%$&$8\times10^{-5}$&43.6&1956&71.7&78.4&62.8&33.6&70.2&-1.56\% \\
  $50\%-70\%$&$8\times10^{-5}$&45.2&1930&71.7&79.4&63.2&35.8&73.5&+0.51\% \\
  \rowcolor{metabg}$60\%-80\%$&$8\times10^{-5}$&44.8&1981&71.9&78.4&63.9&35.3&73.4&\textbf{+0.58\%} \\
\bottomrule
\end{NiceTabular}}
\end{table*}
\subsection{Impact of IRA in Different Transformer Layers } 
Previous studies~\cite{skean2025layer,alain2016understanding,jiang2025devils} have shown that different transformer layers play distinct roles in representation learning. In particular, mid-depth layers progressively compress embeddings and remove noise, while later layers focus more on language-aligned reasoning and output decoding. Prior analyses further indicate that entropy decreases across layers during training, suggesting a gradual consolidation of information.

Motivated by these observations, we investigate where the proposed noisy Attention should be applied within the LLM decode of the VLM system. For InternVL2.5-8B, whose language model has 32 layers, we apply IRA to different layer ranges and train the model for 10k steps.

As shown in Table~\ref{tab:layerstudy}, applying IRA to a contiguous mid-to-late subset of layers (60\%–80\%) achieves the best overall performance in our study, yielding improvements of +0.58\%, respectively. These configurations consistently improve performance on multimodal reasoning benchmarks such as MME, VQA$^{text}$, and CVBench, while maintaining competitive results on spatial tasks like EmbSpatial. This suggests that mid-to-late layers are the most effective stage for regulating visual information, where representations are sufficiently abstract to benefit from compression but still retain critical visual grounding signals.
In contrast, applying IRA across all layers (0\%–100\%) or over a broad range (20\%–80\%) degrades performance. This indicates that over-regularization across the entire network can excessively regularize visual information, hindering both reasoning and generalization. Similarly, applying IRA to earlier layers (20\%–40\%) yields only limited gains, since these layers primarily encode low-level visual features and lack strong cross-modal interactions.

Overall, these results suggest that mid-to-late transformer layers are the critical stage for integrating and refining visual representations, and that applying regularization in this region achieves the best trade-off between visual compression and reasoning capability.

\subsection{IRA Reduces Visual Attention Sink}
\begin{wraptable}{r}{5cm}
\centering 
\caption{Performance and attention sink ratio analysis.}\label{tab:sinkratio}
\scalebox{0.8}{
\begin{NiceTabular}{@{}lcc@{}}
\toprule
Method                   & Acc. $\uparrow$ & Sink Ratio $\downarrow$ \\ \midrule
InternVL2.5-Pt &     48.8        & 96.9\%       \\
+ SFT                    &   53.8          & 46.9\%       \\
+ IRA                    &    \textbf{54.4}         & \textbf{40.6\%}      \\ \bottomrule
\end{NiceTabular}}
\end{wraptable}
To quantify the model's dependence on visual inputs, we provide a layer-wise analysis of the visual attention distribution and sink ratio. Specifically, we first extract the attention weights at layer $l$ as $\mathcal{A}^{(\ell)} \in \mathbb{R}^{\mathcal{H} \times \mathcal{T} \times \mathcal{S}}$ assigned by teacher-forced output tokens to each input visual tokens, where $\mathcal{T}$ and $\mathcal{S}$ are the number of output and visual tokens, and $\mathcal{H}$ is the number of heads. For each transformer layer, the aggregated visual attention map is computed as $A^{(\ell)} = \frac{1}{\mathcal{T} \cdot \mathcal{H}} \sum_{i=1}^{\mathcal{T}} \sum_{n=1}^{\mathcal{H}} \mathcal{A}^{(\ell)}_{i, n}$. 
For all the visual tokens in the sequence, an attention sink~\cite{sun2026spike} exists when there are tokens that receive more than $\epsilon$ average attention, i.e., ${\scriptstyle s^{(\ell)}_\epsilon=\underset{1\leq k\leq \mathcal{S}}{\text{max}}A_k^{(\ell)}> \epsilon }$. Finally, we report the model-level sink ratio by averaging $s^{(\ell)}_\epsilon$ over all the layers. In our experiments, we use $\epsilon=0.15$ consistently. As shown in Table~\ref{tab:sinkratio}, IRA suppresses the sink ratio and improves overall model performance.

\section{Conclusion}
In this work, we address the problem of visual representation learning in vision–language models and identify the lack of explicit control over visual embeddings as a central cause of low interpretability, robustness, and catastrophic forgetting. We introduce Information-Regularized Attention (IRA), a stochastic attention mechanism that injects data-dependent noise into hidden states, enabling adaptive and principled regularization during end-to-end training. Through extensive analysis, we show that IRA not only improves downstream performance but also fundamentally reshapes the geometry of learned representations, yielding smoother curvature trajectories and mitigating attention sink behavior. These findings suggest that adaptively regularizing internal representations is critical for stabilizing multimodal reasoning. More broadly, our results highlight that attention should be viewed not merely as a weighting mechanism but as a metric for evaluating the internal representation. We hope this perspective opens new avenues for designing robust and interpretable multimodal systems.\\
\noindent\textbf{Limitation} 
Due to resource constraints, we apply the proposed methods to models up to 8B parameters, but we expect the conclusions to hold for larger models with more parameters, such as 13B, 26B, and 73B. Additionally, we believe that an IRA can serve as a general architecture for robust representation learning during the pre-training stage of LLMs and VLMs. We leave this to future study.

\clearpage
\newpage
\bibliographystyle{assets/plainnat}
\bibliography{paper}

\begin{thebibliography}{80}
\providecommand{\natexlab}[1]{#1}
\providecommand{\url}[1]{\texttt{#1}}
\expandafter\ifx\csname urlstyle\endcsname\relax
  \providecommand{\doi}[1]{doi: #1}\else
  \providecommand{\doi}{doi: \begingroup \urlstyle{rm}\Url}\fi

\bibitem[Alain and Bengio(2016)]{alain2016understanding}
Guillaume Alain and Yoshua Bengio.
\newblock Understanding intermediate layers using linear classifier probes.
\newblock \emph{arXiv preprint arXiv:1610.01644}, 2016.

\bibitem[Alayrac et~al.(2022)Alayrac, Donahue, Luc, Miech, Barr, Hasson, Lenc, Mensch, Millican, Reynolds, et~al.]{alayrac2022flamingo}
Jean-Baptiste Alayrac, Jeff Donahue, Pauline Luc, Antoine Miech, Iain Barr, Yana Hasson, Karel Lenc, Arthur Mensch, Katherine Millican, Malcolm Reynolds, et~al.
\newblock Flamingo: a visual language model for few-shot learning.
\newblock \emph{Advances in neural information processing systems}, 2022.

\bibitem[Alemi et~al.(2017)Alemi, Fischer, Dillon, and Murphy]{alemi2017deep}
Alexander~A. Alemi, Ian Fischer, Joshua~V. Dillon, and Kevin Murphy.
\newblock Deep variational information bottleneck.
\newblock In \emph{International Conference on Learning Representations}, 2017.

\bibitem[Bai et~al.(2023)Bai, Bai, Chu, Cui, Dang, Deng, Fan, Ge, Han, Huang, et~al.]{bai2023qwen}
Jinze Bai, Shuai Bai, Yunfei Chu, Zeyu Cui, Kai Dang, Xiaodong Deng, Yang Fan, Wenbin Ge, Yu~Han, Fei Huang, et~al.
\newblock Qwen technical report.
\newblock \emph{arXiv preprint arXiv:2309.16609}, 2023.

\bibitem[Barbero et~al.(2025)Barbero, Arroyo, Gu, Perivolaropoulos, Bronstein, Veli{\v{c}}kovi{\'c}, and Pascanu]{barbero2025llms}
Federico Barbero, Alvaro Arroyo, Xiangming Gu, Christos Perivolaropoulos, Michael Bronstein, Petar Veli{\v{c}}kovi{\'c}, and Razvan Pascanu.
\newblock Why do llms attend to the first token?
\newblock \emph{arXiv preprint arXiv:2504.02732}, 2025.

\bibitem[Belrose et~al.(2023)Belrose, Furman, Smith, Halawi, Ostrovsky, McKinney, Biderman, and Steinhardt]{belrose2023eliciting}
Nora Belrose, Zach Furman, Logan Smith, Danny Halawi, Igor Ostrovsky, Lev McKinney, Stella Biderman, and Jacob Steinhardt.
\newblock Eliciting latent predictions from transformers with the tuned lens.
\newblock \emph{arXiv preprint arXiv:2303.08112}, 2023.

\bibitem[Bowman et~al.(2016)Bowman, Vilnis, Vinyals, Dai, Jozefowicz, and Bengio]{bowman2016generating}
Samuel Bowman, Luke Vilnis, Oriol Vinyals, Andrew Dai, Rafal Jozefowicz, and Samy Bengio.
\newblock Generating sentences from a continuous space.
\newblock In \emph{Proceedings of the 20th SIGNLL conference on computational natural language learning}, pages 10--21, 2016.

\bibitem[Chen et~al.(2024{\natexlab{a}})Chen, Li, Dong, Zhang, Zang, Chen, Duan, Wang, Qiao, Lin, and Zhao]{mmstar}
Lin Chen, Jinsong Li, Xiaoyi Dong, Pan Zhang, Yuhang Zang, Zehui Chen, Haodong Duan, Jiaqi Wang, Yu~Qiao, Dahua Lin, and Feng Zhao.
\newblock Are we on the right way for evaluating large vision-language models?
\newblock In \emph{The Thirty-eighth Annual Conference on Neural Information Processing Systems}, 2024{\natexlab{a}}.

\bibitem[Chen et~al.(2020)Chen, Hou, Cui, Che, Liu, and Yu]{chen-etal-2020-recall}
Sanyuan Chen, Yutai Hou, Yiming Cui, Wanxiang Che, Ting Liu, and Xiangzhan Yu.
\newblock Recall and learn: Fine-tuning deep pretrained language models with less forgetting.
\newblock In \emph{Proceedings of the 2020 Conference on Empirical Methods in Natural Language Processing (EMNLP)}, 2020.

\bibitem[Chen et~al.(2024{\natexlab{b}})Chen, Wang, Tian, Ye, Gao, Cui, Tong, Hu, Luo, Ma, et~al.]{internvl}
Zhe Chen, Weiyun Wang, Hao Tian, Shenglong Ye, Zhangwei Gao, Erfei Cui, Wenwen Tong, Kongzhi Hu, Jiapeng Luo, Zheng Ma, et~al.
\newblock How far are we to gpt-4v? closing the gap to commercial multimodal models with open-source suites.
\newblock \emph{Science China Information Sciences}, 2024{\natexlab{b}}.

\bibitem[de~Llano et~al.(2026)de~Llano, Arroyo, Barbero, Dong, Bronstein, LeCun, and Shwartz-Ziv]{queipo-de-llano2026attention}
Enrique~Queipo de~Llano, Alvaro Arroyo, Federico Barbero, Xiaowen Dong, Michael~M. Bronstein, Yann LeCun, and Ravid Shwartz-Ziv.
\newblock Attention sinks and compression valleys in {LLM}s are two sides of the same coin.
\newblock In \emph{The Fourteenth International Conference on Learning Representations}, 2026.

\bibitem[Dong et~al.(2021)Dong, Luu, Lin, Yan, and Zhang]{dong2021should}
Xinshuai Dong, Anh~Tuan Luu, Min Lin, Shuicheng Yan, and Hanwang Zhang.
\newblock How should pre-trained language models be fine-tuned towards adversarial robustness?
\newblock \emph{Advances in Neural Information Processing Systems}, 2021.

\bibitem[Du et~al.(2024)Du, Wu, Li, Huang, and Wei]{du2024embspatial}
Mengfei Du, Binhao Wu, Zejun Li, Xuan-Jing Huang, and Zhongyu Wei.
\newblock Embspatial-bench: Benchmarking spatial understanding for embodied tasks with large vision-language models.
\newblock In \emph{Proceedings of the 62nd Annual Meeting of the Association for Computational Linguistics (Volume 2: Short Papers)}, 2024.

\bibitem[Dziugaite and Roy(2018)]{dziugaite2018data}
Gintare~Karolina Dziugaite and Daniel~M Roy.
\newblock Data-dependent pac-bayes priors via differential privacy.
\newblock \emph{Advances in neural information processing systems}, 31, 2018.

\bibitem[Fan et~al.(2020)Fan, Zhang, Chen, and Zhou]{fan2020bayesian}
Xinjie Fan, Shujian Zhang, Bo~Chen, and Mingyuan Zhou.
\newblock Bayesian attention modules.
\newblock \emph{Advances in Neural Information Processing Systems}, 2020.

\bibitem[Fu et~al.(2023)Fu, Chen, Shen, Qin, Zhang, Lin, Qiu, Lin, Yang, Zheng, Li, Sun, and Ji]{Fu2023MMEAC}
Chaoyou Fu, Peixian Chen, Yunhang Shen, Yulei Qin, Mengdan Zhang, Xu~Lin, Zhenyu Qiu, Wei Lin, Jinrui Yang, Xiawu Zheng, Ke~Li, Xing Sun, and Rongrong Ji.
\newblock Mme: A comprehensive evaluation benchmark for multimodal large language models.
\newblock \emph{ArXiv}, 2023.

\bibitem[Fu et~al.(2024)Fu, Hu, Li, Feng, Wang, Lin, Roth, Smith, Ma, and Krishna]{fu2024blink}
Xingyu Fu, Yushi Hu, Bangzheng Li, Yu~Feng, Haoyu Wang, Xudong Lin, Dan Roth, Noah~A Smith, Wei-Chiu Ma, and Ranjay Krishna.
\newblock Blink: Multimodal large language models can see but not perceive.
\newblock In \emph{European Conference on Computer Vision}, 2024.

\bibitem[Gan et~al.(2022)Gan, Li, Li, Wang, Liu, Gao, et~al.]{gan2022vision}
Zhe Gan, Linjie Li, Chunyuan Li, Lijuan Wang, Zicheng Liu, Jianfeng Gao, et~al.
\newblock Vision-language pre-training: Basics, recent advances, and future trends.
\newblock \emph{Foundations and Trends{\textregistered} in Computer Graphics and Vision}, 2022.

\bibitem[Gu et~al.(2025)Gu, Pang, Du, Liu, Zhang, Du, Wang, and Lin]{gu2024attention}
Xiangming Gu, Tianyu Pang, Chao Du, Qian Liu, Fengzhuo Zhang, Cunxiao Du, Ye~Wang, and Min Lin.
\newblock When attention sink emerges in language models: An empirical view.
\newblock In \emph{ICLR}, 2025.

\bibitem[Guan et~al.(2024)Guan, Liu, Wu, Xian, Li, Liu, Wang, Chen, Huang, Yacoob, et~al.]{guan2024hallusionbench}
Tianrui Guan, Fuxiao Liu, Xiyang Wu, Ruiqi Xian, Zongxia Li, Xiaoyu Liu, Xijun Wang, Lichang Chen, Furong Huang, Yaser Yacoob, et~al.
\newblock Hallusionbench: an advanced diagnostic suite for entangled language hallucination and visual illusion in large vision-language models.
\newblock In \emph{Proceedings of the IEEE/CVF Conference on Computer Vision and Pattern Recognition}, 2024.

\bibitem[H{\'e}naff et~al.(2019)H{\'e}naff, Goris, and Simoncelli]{henaff2019perceptual}
Olivier~J H{\'e}naff, Robbe~LT Goris, and Eero~P Simoncelli.
\newblock Perceptual straightening of natural videos.
\newblock \emph{Nature neuroscience}, 22\penalty0 (6):\penalty0 984--991, 2019.

\bibitem[Hong et~al.(2025)Hong, Kim, Jeon, and Lee]{hong2025comprehensive}
Jung-Ho Hong, Ho-Joong Kim, Kyu-Sung Jeon, and Seong-Whan Lee.
\newblock Comprehensive information bottleneck for unveiling universal attribution to interpret vision transformers.
\newblock In \emph{Proceedings of the Computer Vision and Pattern Recognition Conference}, 2025.

\bibitem[Hosseini and Fedorenko(2023)]{hosseini2023large}
Eghbal~A. Hosseini and Evelina Fedorenko.
\newblock Large language models implicitly learn to straighten neural sentence trajectories to construct a predictive representation of natural language.
\newblock In \emph{Thirty-seventh Conference on Neural Information Processing Systems}, 2023.
\newblock \url{https://openreview.net/forum?id=h3lTrt4Ftb}.

\bibitem[Jiang et~al.(2024)Jiang, Liu, and Zheng]{jiang2024correlation}
Jingjing Jiang, Ziyi Liu, and Nanning Zheng.
\newblock Correlation information bottleneck: Towards adapting pretrained multimodal models for robust visual question answering.
\newblock \emph{International Journal of Computer Vision}, 2024.

\bibitem[Jiang et~al.(2025)Jiang, Chen, Zhu, Luo, Shen, and Yang]{jiang2025devils}
Zhangqi Jiang, Junkai Chen, Beier Zhu, Tingjin Luo, Yankun Shen, and Xu~Yang.
\newblock Devils in middle layers of large vision-language models: Interpreting, detecting and mitigating object hallucinations via attention lens.
\newblock In \emph{Proceedings of the Computer Vision and Pattern Recognition Conference}, 2025.

\bibitem[Kang et~al.(2025)Kang, Kim, Kim, and Hwang]{kang2025see}
Seil Kang, Jinyeong Kim, Junhyeok Kim, and Seong~Jae Hwang.
\newblock See what you are told: Visual attention sink in large multimodal models.
\newblock \emph{arXiv preprint arXiv:2503.03321}, 2025.

\bibitem[Li et~al.(2026)Li, Ni, Qu, Miao, Yang, Fu, Chen, and Cheng]{li2026reinforced}
Bangzheng Li, Jianmo Ni, Chen Qu, Ian Miao, Liu Yang, Xingyu Fu, Muhao Chen, and Derek~Zhiyuan Cheng.
\newblock Reinforced attention learning.
\newblock \emph{arXiv preprint arXiv:2602.04884}, 2026.

\bibitem[Li et~al.(2024{\natexlab{a}})Li, Zhang, Guo, Zhang, Li, Zhang, Zhang, Zhang, Li, Liu, et~al.]{li2024llava}
Bo~Li, Yuanhan Zhang, Dong Guo, Renrui Zhang, Feng Li, Hao Zhang, Kaichen Zhang, Peiyuan Zhang, Yanwei Li, Ziwei Liu, et~al.
\newblock Llava-onevision: Easy visual task transfer.
\newblock \emph{arXiv preprint arXiv:2408.03326}, 2024{\natexlab{a}}.

\bibitem[Li et~al.(2023{\natexlab{a}})Li, Li, Savarese, and Hoi]{li2023blip}
Junnan Li, Dongxu Li, Silvio Savarese, and Steven Hoi.
\newblock Blip-2: Bootstrapping language-image pre-training with frozen image encoders and large language models.
\newblock In \emph{International conference on machine learning}. PMLR, 2023{\natexlab{a}}.

\bibitem[Li et~al.(2024{\natexlab{b}})Li, Wang, He, Li, Wang, Liu, Wang, Xu, Chen, Luo, et~al.]{li2024mvbench}
Kunchang Li, Yali Wang, Yinan He, Yizhuo Li, Yi~Wang, Yi~Liu, Zun Wang, Jilan Xu, Guo Chen, Ping Luo, et~al.
\newblock Mvbench: A comprehensive multi-modal video understanding benchmark.
\newblock In \emph{Proceedings of the IEEE/CVF Conference on Computer Vision and Pattern Recognition}, pages 22195--22206, 2024{\natexlab{b}}.

\bibitem[Li et~al.(2023{\natexlab{b}})Li, Du, Zhou, Wang, Zhao, and Wen]{li2023evaluating}
Yifan Li, Yifan Du, Kun Zhou, Jinpeng Wang, Wayne~Xin Zhao, and Ji-Rong Wen.
\newblock Evaluating object hallucination in large vision-language models.
\newblock \emph{arXiv preprint arXiv:2305.10355}, 2023{\natexlab{b}}.

\bibitem[Li et~al.(2023{\natexlab{c}})Li, Du, Zhou, Wang, Zhao, and Wen]{li-etal-2023-evaluating}
Yifan Li, Yifan Du, Kun Zhou, Jinpeng Wang, Xin Zhao, and Ji-Rong Wen.
\newblock Evaluating object hallucination in large vision-language models.
\newblock In \emph{Proceedings of the 2023 Conference on Empirical Methods in Natural Language Processing}, 2023{\natexlab{c}}.

\bibitem[Li et~al.(2020)Li, Wang, Chen, Utiyama, Sumita, Zhang, and Zhao]{Li2020Data-dependent}
Zuchao Li, Rui Wang, Kehai Chen, Masso Utiyama, Eiichiro Sumita, Zhuosheng Zhang, and Hai Zhao.
\newblock Data-dependent gaussian prior objective for language generation.
\newblock In \emph{International Conference on Learning Representations}, 2020.

\bibitem[Liu et~al.(2023)Liu, Li, Wu, and Lee]{liu2023visual}
Haotian Liu, Chunyuan Li, Qingyang Wu, and Yong~Jae Lee.
\newblock Visual instruction tuning.
\newblock \emph{Advances in neural information processing systems}, 2023.

\bibitem[Lu et~al.(2019)Lu, Batra, Parikh, and Lee]{lu2019vilbert}
Jiasen Lu, Dhruv Batra, Devi Parikh, and Stefan Lee.
\newblock Vilbert: Pretraining task-agnostic visiolinguistic representations for vision-and-language tasks.
\newblock \emph{Advances in neural information processing systems}, 2019.

\bibitem[Mahajan et~al.(2025)Mahajan, Le, Park, Farhadzadeh, Hayat, and Porikli]{mahajan2025attention}
Shweta Mahajan, Hoang Le, Hyojin Park, Farzad Farhadzadeh, Munawar Hayat, and Fatih Porikli.
\newblock Attention guided alignment in efficient vision-language models.
\newblock \emph{arXiv preprint arXiv:2511.17793}, 2025.

\bibitem[Mao et~al.(2016)Mao, Huang, Toshev, Camburu, Yuille, and Murphy]{mao2016generation}
Junhua Mao, Jonathan Huang, Alexander Toshev, Oana Camburu, Alan~L Yuille, and Kevin Murphy.
\newblock Generation and comprehension of unambiguous object descriptions.
\newblock In \emph{Proceedings of the IEEE conference on computer vision and pattern recognition}, pages 11--20, 2016.

\bibitem[Marino et~al.(2019)Marino, Rastegari, Farhadi, and Mottaghi]{marino2019ok}
Kenneth Marino, Mohammad Rastegari, Ali Farhadi, and Roozbeh Mottaghi.
\newblock Ok-vqa: A visual question answering benchmark requiring external knowledge.
\newblock In \emph{Proceedings of the IEEE/cvf conference on computer vision and pattern recognition}, 2019.

\bibitem[Masry et~al.(2022)Masry, Do, Tan, Joty, and Hoque]{masry-etal-2022-chartqa}
Ahmed Masry, Xuan~Long Do, Jia~Qing Tan, Shafiq Joty, and Enamul Hoque.
\newblock {C}hart{QA}: A benchmark for question answering about charts with visual and logical reasoning.
\newblock In \emph{Findings of the Association for Computational Linguistics: ACL 2022}, 2022.

\bibitem[Mathew et~al.(2021)Mathew, Karatzas, and Jawahar]{mathew2021docvqa}
Minesh Mathew, Dimosthenis Karatzas, and CV~Jawahar.
\newblock Docvqa: A dataset for vqa on document images.
\newblock In \emph{Proceedings of the IEEE/CVF winter conference on applications of computer vision}, 2021.

\bibitem[Noh et~al.(2017)Noh, You, Mun, and Han]{noh2017regularizing}
Hyeonwoo Noh, Tackgeun You, Jonghwan Mun, and Bohyung Han.
\newblock Regularizing deep neural networks by noise: Its interpretation and optimization.
\newblock \emph{Advances in neural information processing systems}, 30, 2017.

\bibitem[Ouyang et~al.(2022)Ouyang, Wu, Jiang, Almeida, Wainwright, Mishkin, Zhang, Agarwal, Slama, Ray, et~al.]{ouyang2022training}
Long Ouyang, Jeffrey Wu, Xu~Jiang, Diogo Almeida, Carroll Wainwright, Pamela Mishkin, Chong Zhang, Sandhini Agarwal, Katarina Slama, Alex Ray, et~al.
\newblock Training language models to follow instructions with human feedback.
\newblock \emph{Advances in neural information processing systems}, 2022.

\bibitem[Peng et~al.(2025)Peng, Yang, Jiang, and Tian]{Peng_2025_ICCV}
Shangpin Peng, Senqiao Yang, Li~Jiang, and Zhuotao Tian.
\newblock Mitigating object hallucinations via sentence-level early intervention.
\newblock In \emph{Proceedings of the IEEE/CVF International Conference on Computer Vision (ICCV)}, October 2025.

\bibitem[Poole et~al.(2014)Poole, Sohl-Dickstein, and Ganguli]{poole2014analyzing}
Ben Poole, Jascha Sohl-Dickstein, and Surya Ganguli.
\newblock Analyzing noise in autoencoders and deep networks.
\newblock \emph{arXiv preprint arXiv:1406.1831}, 2014.

\bibitem[Qiu et~al.(2025)Qiu, Wang, Zheng, Huang, Wen, Yang, Men, Yu, Huang, Huang, Liu, Zhou, and Lin]{qiu2025gated}
Zihan Qiu, Zekun Wang, Bo~Zheng, Zeyu Huang, Kaiyue Wen, Songlin Yang, Rui Men, Le~Yu, Fei Huang, Suozhi Huang, Dayiheng Liu, Jingren Zhou, and Junyang Lin.
\newblock Gated attention for large language models: Non-linearity, sparsity, and attention-sink-free.
\newblock In \emph{The Thirty-ninth Annual Conference on Neural Information Processing Systems}, 2025.

\bibitem[Rafailov et~al.(2023)Rafailov, Sharma, Mitchell, Manning, Ermon, and Finn]{rafailov2023direct}
Rafael Rafailov, Archit Sharma, Eric Mitchell, Christopher~D Manning, Stefano Ermon, and Chelsea Finn.
\newblock Direct preference optimization: Your language model is secretly a reward model.
\newblock \emph{Advances in neural information processing systems}, 2023.

\bibitem[Raghu et~al.(2017)Raghu, Gilmer, Yosinski, and Sohl-Dickstein]{raghu2017svcca}
Maithra Raghu, Justin Gilmer, Jason Yosinski, and Jascha Sohl-Dickstein.
\newblock Svcca: Singular vector canonical correlation analysis for deep learning dynamics and interpretability.
\newblock \emph{Advances in neural information processing systems}, 30, 2017.

\bibitem[Rahmanzadehgervi et~al.(2024)Rahmanzadehgervi, Bolton, Taesiri, and Nguyen]{rahmanzadehgervi2024vision}
Pooyan Rahmanzadehgervi, Logan Bolton, Mohammad~Reza Taesiri, and Anh~Totti Nguyen.
\newblock Vision language models are blind: Failing to translate detailed visual features into words.
\newblock \emph{arXiv preprint arXiv:2407.06581}, 2024.

\bibitem[Rohrbach et~al.(2018)Rohrbach, Hendricks, Burns, Darrell, and Saenko]{rohrbach-etal-2018-object}
Anna Rohrbach, Lisa~Anne Hendricks, Kaylee Burns, Trevor Darrell, and Kate Saenko.
\newblock Object hallucination in image captioning.
\newblock In \emph{Proceedings of the 2018 Conference on Empirical Methods in Natural Language Processing}, 2018.

\bibitem[Schulman et~al.(2017)Schulman, Wolski, Dhariwal, Radford, and Klimov]{schulman2017proximal}
John Schulman, Filip Wolski, Prafulla Dhariwal, Alec Radford, and Oleg Klimov.
\newblock Proximal policy optimization algorithms.
\newblock \emph{arXiv preprint arXiv:1707.06347}, 2017.

\bibitem[Shao et~al.(2024)Shao, Wang, Zhu, Xu, Song, Zhang, Li, Wu, and Guo]{Shao2024DeepSeekMathPT}
Zhihong Shao, Peiyi Wang, Qihao Zhu, Runxin Xu, Jun-Mei Song, Mingchuan Zhang, Y.~K. Li, Yu~Wu, and Daya Guo.
\newblock Deepseekmath: Pushing the limits of mathematical reasoning in open language models.
\newblock \emph{ArXiv}, 2024.

\bibitem[Singh et~al.(2019)Singh, Natarajan, Shah, Jiang, Chen, Batra, Parikh, and Rohrbach]{textvqa}
Amanpreet Singh, Vivek Natarajan, Meet Shah, Yu~Jiang, Xinlei Chen, Dhruv Batra, Devi Parikh, and Marcus Rohrbach.
\newblock Towards vqa models that can read.
\newblock In \emph{Proceedings of the IEEE/CVF conference on computer vision and pattern recognition}, pages 8317--8326, 2019.

\bibitem[Skean et~al.(2025)Skean, Arefin, Zhao, Patel, Naghiyev, LeCun, and Shwartz-Ziv]{skean2025layer}
Oscar Skean, Md~Rifat Arefin, Dan Zhao, Niket Patel, Jalal Naghiyev, Yann LeCun, and Ravid Shwartz-Ziv.
\newblock Layer by layer: Uncovering hidden representations in language models.
\newblock \emph{arXiv preprint arXiv:2502.02013}, 2025.

\bibitem[S{\o}nderby et~al.(2016)S{\o}nderby, Raiko, Maal{\o}e, S{\o}nderby, and Winther]{sonderby2016ladder}
Casper~Kaae S{\o}nderby, Tapani Raiko, Lars Maal{\o}e, S{\o}ren~Kaae S{\o}nderby, and Ole Winther.
\newblock Ladder variational autoencoders.
\newblock \emph{Advances in neural information processing systems}, 2016.

\bibitem[Srivastava et~al.(2014)Srivastava, Hinton, Krizhevsky, Sutskever, and Salakhutdinov]{10.5555/2627435.2670313}
Nitish Srivastava, Geoffrey Hinton, Alex Krizhevsky, Ilya Sutskever, and Ruslan Salakhutdinov.
\newblock Dropout: a simple way to prevent neural networks from overfitting.
\newblock \emph{J. Mach. Learn. Res.}, 2014.

\bibitem[Steinberg and Gal(2026)]{steinberg2026vision}
Jonathan Steinberg and Oren Gal.
\newblock Where vision becomes text: Locating the ocr routing bottleneck in vision-language models.
\newblock \emph{arXiv preprint arXiv:2602.22918}, 2026.

\bibitem[Sun et~al.(2024{\natexlab{a}})Sun, Qin, Fu, Wang, and Tao]{sun-etal-2024-self}
Guohao Sun, Can Qin, Huazhu Fu, Linwei Wang, and Zhiqiang Tao.
\newblock Self-training large language and vision assistant for medical question answering.
\newblock In \emph{Proceedings of the 2024 Conference on Empirical Methods in Natural Language Processing}, November 2024{\natexlab{a}}.

\bibitem[Sun et~al.(2024{\natexlab{b}})Sun, Qin, Wang, Chen, Xu, and Tao]{sun2024sq}
Guohao Sun, Can Qin, Jiamian Wang, Zeyuan Chen, Ran Xu, and Zhiqiang Tao.
\newblock Sq-llava: Self-questioning for large vision-language assistant.
\newblock In \emph{European Conference on Computer Vision}, pages 156--172, 2024{\natexlab{b}}.

\bibitem[Sun et~al.(2025{\natexlab{a}})Sun, Hua, Wang, Luo, Dianat, RABBANI, Rao, and Tao]{NEURIPS2025_95c6ae3f}
Guohao Sun, Hang Hua, Jian Wang, Jiebo Luo, Sohail Dianat, MAJID RABBANI, Raghuveer Rao, and Zhiqiang Tao.
\newblock Latent chain-of-thought for visual reasoning.
\newblock In D.~Belgrave, C.~Zhang, H.~Lin, R.~Pascanu, P.~Koniusz, M.~Ghassemi, and N.~Chen, editors, \emph{Advances in Neural Information Processing Systems}, 2025{\natexlab{a}}.

\bibitem[Sun et~al.(2025{\natexlab{b}})Sun, Qin, Feng, Chen, Xu, Dianat, Rabbani, Rao, and Tao]{Sun_2025_ICCV}
Guohao Sun, Can Qin, Yihao Feng, Zeyuan Chen, Ran Xu, Sohail Dianat, Majid Rabbani, Raghuveer Rao, and Zhiqiang Tao.
\newblock Structured policy optimization: Enhance large vision-language model via self-referenced dialogue.
\newblock In \emph{Proceedings of the IEEE/CVF International Conference on Computer Vision (ICCV)}, 2025{\natexlab{b}}.

\bibitem[Sun et~al.(2026{\natexlab{a}})Sun, Wang, Ma, Xie, Cheng, Tao, and Wang]{Sun_2026_CVPR}
Guohao Sun, Yufei Wang, Sizhuo Ma, Yuege Xie, Yuting Cheng, Zhiqiang Tao, and Jian Wang.
\newblock If-prune: Information-flow guided token pruning for efficient vision-language models.
\newblock In \emph{Proceedings of the IEEE/CVF Conference on Computer Vision and Pattern Recognition (CVPR)}, pages 3522--3531, June 2026{\natexlab{a}}.

\bibitem[Sun et~al.(2024{\natexlab{c}})Sun, Chen, Kolter, and Liu]{sun2024massive}
Mingjie Sun, Xinlei Chen, J~Zico Kolter, and Zhuang Liu.
\newblock Massive activations in large language models.
\newblock \emph{arXiv preprint arXiv:2402.17762}, 2024{\natexlab{c}}.

\bibitem[Sun et~al.(2026{\natexlab{b}})Sun, Canziani, LeCun, and Zhu]{sun2026spike}
Shangwen Sun, Alfredo Canziani, Yann LeCun, and Jiachen Zhu.
\newblock The spike, the sparse and the sink: Anatomy of massive activations and attention sinks.
\newblock \emph{arXiv preprint arXiv:2603.05498}, 2026{\natexlab{b}}.

\bibitem[Tishby et~al.(2000)Tishby, Pereira, and Bialek]{tishby2000information}
Naftali Tishby, Fernando~C Pereira, and William Bialek.
\newblock The information bottleneck method.
\newblock \emph{arXiv preprint physics/0004057}, 2000.

\bibitem[Vaswani et~al.(2017)Vaswani, Shazeer, Parmar, Uszkoreit, Jones, Gomez, Kaiser, and Polosukhin]{NIPS2017_3f5ee243}
Ashish Vaswani, Noam Shazeer, Niki Parmar, Jakob Uszkoreit, Llion Jones, Aidan~N Gomez, \L~ukasz Kaiser, and Illia Polosukhin.
\newblock Attention is all you need.
\newblock In I.~Guyon, U.~Von Luxburg, S.~Bengio, H.~Wallach, R.~Fergus, S.~Vishwanathan, and R.~Garnett, editors, \emph{Advances in Neural Information Processing Systems}. Curran Associates, Inc., 2017.

\bibitem[Vo et~al.(2025)Vo, Nguyen, Taesiri, Dang, Nguyen, and Kim]{vo2025vision}
An~Vo, Khai-Nguyen Nguyen, Mohammad~Reza Taesiri, Vy~Tuong Dang, Anh~Totti Nguyen, and Daeyoung Kim.
\newblock Vision language models are biased.
\newblock \emph{arXiv preprint arXiv:2505.23941}, 2025.

\bibitem[Voita et~al.(2019)Voita, Sennrich, and Titov]{voita2019bottom}
Elena Voita, Rico Sennrich, and Ivan Titov.
\newblock The bottom-up evolution of representations in the transformer: A study with machine translation and language modeling objectives.
\newblock In \emph{Proceedings of the 2019 Conference on Empirical Methods in Natural Language Processing and the 9th International Joint Conference on Natural Language Processing (EMNLP-IJCNLP)}, 2019.

\bibitem[Wang et~al.(2024)Wang, Fu, Huang, Li, Liu, Liu, Ma, Xu, Zhou, Zhang, Yan, Mo, Liu, Lu, Li, Xiao, Chang, Roth, Zhang, Poon, and Chen]{Wang2024MuirBenchAC}
Fei Wang, Xingyu Fu, James~Y. Huang, Zekun Li, Qin Liu, Xiaogeng Liu, Mingyu~Derek Ma, Nan Xu, Wenxuan Zhou, Kai Zhang, Tianyi Yan, Wenjie~Jacky Mo, Hsiang-Hui Liu, Pan Lu, Chunyuan Li, Chaowei Xiao, Kai-Wei Chang, Dan Roth, Sheng Zhang, Hoifung Poon, and Muhao Chen.
\newblock Muirbench: A comprehensive benchmark for robust multi-image understanding.
\newblock \emph{ArXiv}, 2024.

\bibitem[Witten(2020)]{witten2020mini}
Edward Witten.
\newblock A mini-introduction to information theory.
\newblock \emph{La Rivista del Nuovo Cimento}, 43\penalty0 (4):\penalty0 187--227, 2020.

\bibitem[Wu et~al.(2025)Wu, Xiong, Li, Xia, Wang, Wang, Yu, Kim, Rossi, Yao, Shang, and McAuley]{wu-etal-2025-mitigating-visual}
Junda Wu, Yuxin Xiong, Xintong Li, Yu~Xia, Ruoyu Wang, Yu~Wang, Tong Yu, Sungchul Kim, Ryan~A. Rossi, Lina Yao, Jingbo Shang, and Julian McAuley.
\newblock Mitigating visual knowledge forgetting in {MLLM} instruction-tuning via modality-decoupled gradient descent.
\newblock In \emph{Findings of the Association for Computational Linguistics: EMNLP 2025}, November 2025.

\bibitem[x.ai(2024)]{realworldqa}
x.ai.
\newblock Grok-1.5 vision preview.
\newblock Technical report, x.ai, 2024.

\bibitem[Xiao et~al.(2023{\natexlab{a}})Xiao, Lin, Seznec, Wu, Demouth, and Han]{xiao2023smoothquant}
Guangxuan Xiao, Ji~Lin, Mickael Seznec, Hao Wu, Julien Demouth, and Song Han.
\newblock Smoothquant: Accurate and efficient post-training quantization for large language models.
\newblock In \emph{International conference on machine learning}, pages 38087--38099. PMLR, 2023{\natexlab{a}}.

\bibitem[Xiao et~al.(2023{\natexlab{b}})Xiao, Tian, Chen, Han, and Lewis]{xiao2023efficient}
Guangxuan Xiao, Yuandong Tian, Beidi Chen, Song Han, and Mike Lewis.
\newblock Efficient streaming language models with attention sinks.
\newblock \emph{arXiv preprint arXiv:2309.17453}, 2023{\natexlab{b}}.

\bibitem[Yu et~al.(2016)Yu, Poirson, Yang, Berg, and Berg]{refcoco+}
Licheng Yu, Patrick Poirson, Shan Yang, Alexander~C Berg, and Tamara~L Berg.
\newblock Modeling context in referring expressions.
\newblock In \emph{European conference on computer vision}, 2016.

\bibitem[Yue et~al.(2023)Yue, Ni, Zhang, Zheng, Liu, Zhang, Stevens, Jiang, Ren, Sun, Wei, Yu, Yuan, Sun, Yin, Zheng, Yang, Liu, Huang, Sun, Su, and Chen]{Yue2023MMMUAM}
Xiang Yue, Yuansheng Ni, Kai Zhang, Tianyu Zheng, Ruoqi Liu, Ge~Zhang, Samuel Stevens, Dongfu Jiang, Weiming Ren, Yuxuan Sun, Cong Wei, Botao Yu, Ruibin Yuan, Renliang Sun, Ming Yin, Boyuan Zheng, Zhenzhu Yang, Yibo Liu, Wenhao Huang, Huan Sun, Yu~Su, and Wenhu Chen.
\newblock Mmmu: A massive multi-discipline multimodal understanding and reasoning benchmark for expert agi.
\newblock \emph{2024 IEEE/CVF Conference on Computer Vision and Pattern Recognition (CVPR)}, 2023.

\bibitem[Yue et~al.(2025)Yue, Zheng, Ni, Wang, Zhang, Tong, Sun, Yu, Zhang, Sun, et~al.]{yue2025mmmu}
Xiang Yue, Tianyu Zheng, Yuansheng Ni, Yubo Wang, Kai Zhang, Shengbang Tong, Yuxuan Sun, Botao Yu, Ge~Zhang, Huan Sun, et~al.
\newblock Mmmu-pro: A more robust multi-discipline multimodal understanding benchmark.
\newblock In \emph{Proceedings of the 63rd Annual Meeting of the Association for Computational Linguistics (Volume 1: Long Papers)}, pages 15134--15186, 2025.

\bibitem[Zhai et~al.(2023)Zhai, Tong, Li, Cai, Qu, Lee, and Ma]{zhai2023investigating}
Yuexiang Zhai, Shengbang Tong, Xiao Li, Mu~Cai, Qing Qu, Yong~Jae Lee, and Yi~Ma.
\newblock Investigating the catastrophic forgetting in multimodal large language model fine-tuning.
\newblock In \emph{Conference on Parsimony and Learning (Proceedings Track)}, 2023.

\bibitem[Zhang et~al.(2026)Zhang, Wu, Wang, Wang, Lv, Huang, and Zheng]{zhang2026vib}
Feiran Zhang, Yixin Wu, Zhenghua Wang, Xiaohua Wang, Changze Lv, Xuanjing Huang, and Xiaoqing Zheng.
\newblock Vib-probe: Detecting and mitigating hallucinations in vision-language models via variational information bottleneck.
\newblock \emph{arXiv preprint arXiv:2601.05547}, 2026.

\bibitem[Zhang et~al.(2024)Zhang, Zhang, Tian, Fu, Zhang, Wu, Li, Wang, Wen, Zhang, et~al.]{zhang2024mme}
Yi-Fan Zhang, Huanyu Zhang, Haochen Tian, Chaoyou Fu, Shuangqing Zhang, Junfei Wu, Feng Li, Kun Wang, Qingsong Wen, Zhang Zhang, et~al.
\newblock Mme-realworld: Could your multimodal llm challenge high-resolution real-world scenarios that are difficult for humans?
\newblock \emph{arXiv preprint arXiv:2408.13257}, 2024.

\bibitem[Zhao et~al.(2025)Zhao, Zhang, Sun, and Feng]{zhao-etal-2025-mitigating}
Jianfei Zhao, Feng Zhang, Xin Sun, and Chong Feng.
\newblock Mitigating hallucination in large vision-language models through aligning attention distribution to information flow.
\newblock In \emph{Findings of the Association for Computational Linguistics: EMNLP 2025}, 2025.

\end{thebibliography}

\clearpage
\newpage
\beginappendix

\section{Analysis}
\subsection{Correlation Between Model Attention and Prediction}
\begin{figure}[h]
  \centering
  \begin{subfigure}{0.48\linewidth}
  \centering
    \includegraphics[width=\linewidth]{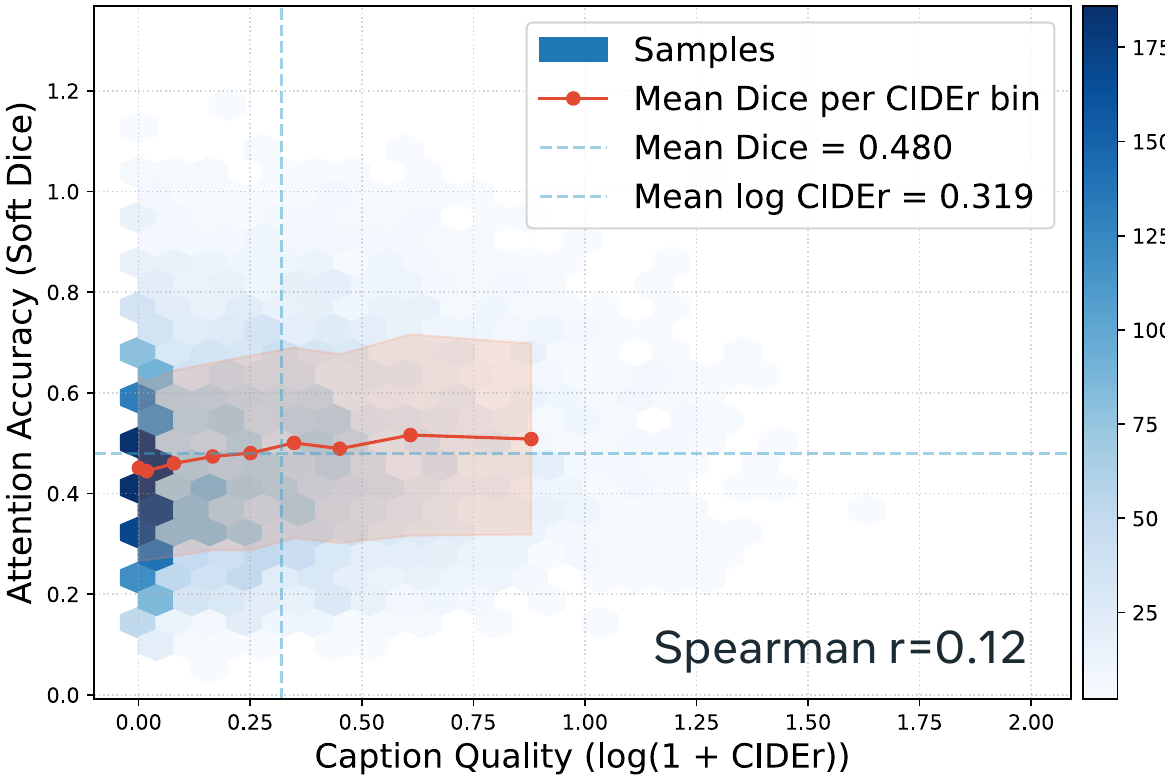}
    \caption{Attention vs. Performance by SFT}
    \label{fig:cor-sft}
  \end{subfigure}
  \hfill
  \begin{subfigure}{0.48\linewidth}
  \centering
    \includegraphics[width=\linewidth]{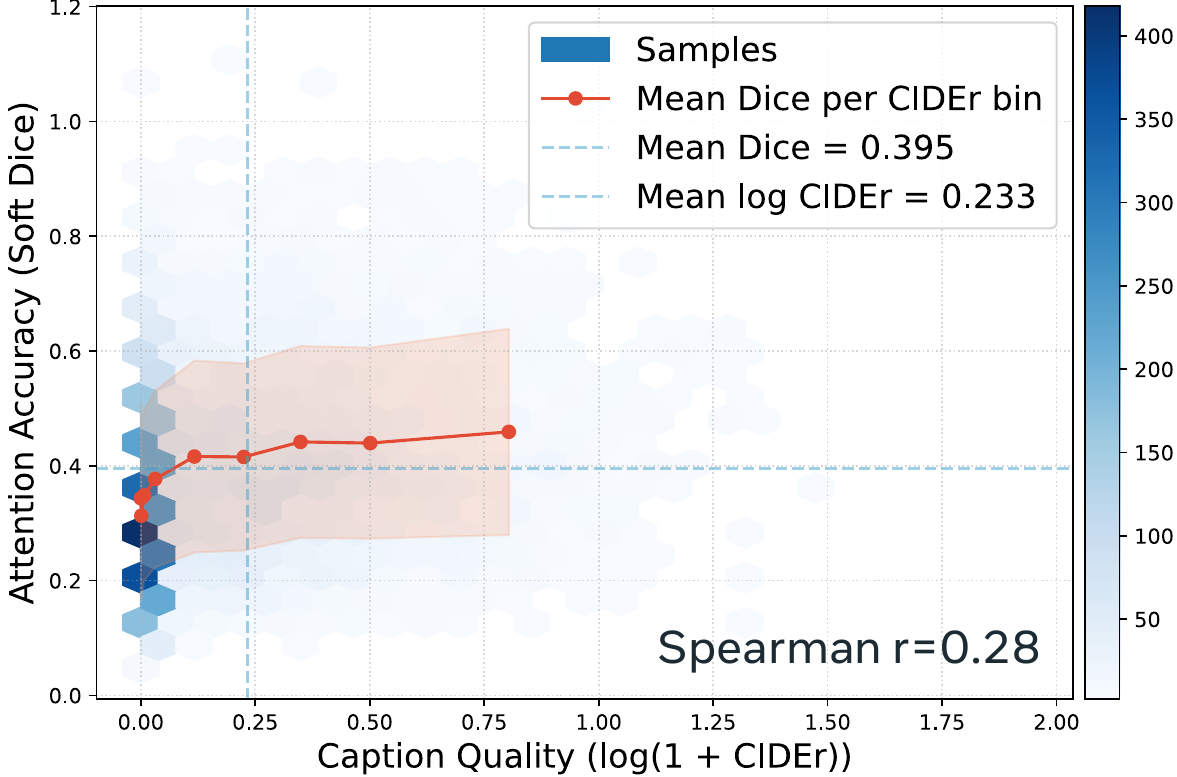}
    \caption{Attention vs. Performance of by IRA}
    \label{fig:cor-ira}
  \end{subfigure} 
  \caption{\label{fig:correlation}Relationship between attention accuracy (soft dice) and VLM performance. The red curve indicates the overall monotonic trend. While SFT and IRA both show positive correlations, IRA has a \underline{higher Spearman} correlation (r=0.28, p=0.0), suggesting that the performance gain is \underline{more strongly associated} with improved attention grounding.}
\end{figure}

To examine the relationship between visual attention accuracy and performance, we conduct experiments on datasets that provide bounding-box annotations indicating the locations of answer-relevant objects. By projecting token-level attention maps into pixel space, we can evaluate the accuracy of attention allocation against the `ground-truth' attention map using the Soft Dice metric. In Fig.~\ref{fig:correlation}, we evaluate InternVL2-8B~\cite{internvl} on the RefCOCOg~\cite{mao2016generation} captioning task and analyze how attention grounding correlates with generation quality. After training with SFT, we have observed a monotonic increase in CIDEr performance with increasing attention accuracy. Given this observation, we hypothesize that improving attention accuracy is crucial for performance gains and that increasing the correlation between them can improve the model's interpretability. Therefore, improving attention behavior is not merely a robustness objective, but a functional requirement for aligning visual perception with language reasoning and grounding model decisions to meaningful visual cues.

\subsection{Catastrophic Forgetting Analysis}\label{appen:forget}
\begin{figure}[t]
  \centering
  \begin{subfigure}{0.45\linewidth}
    \includegraphics[width=0.9\linewidth]{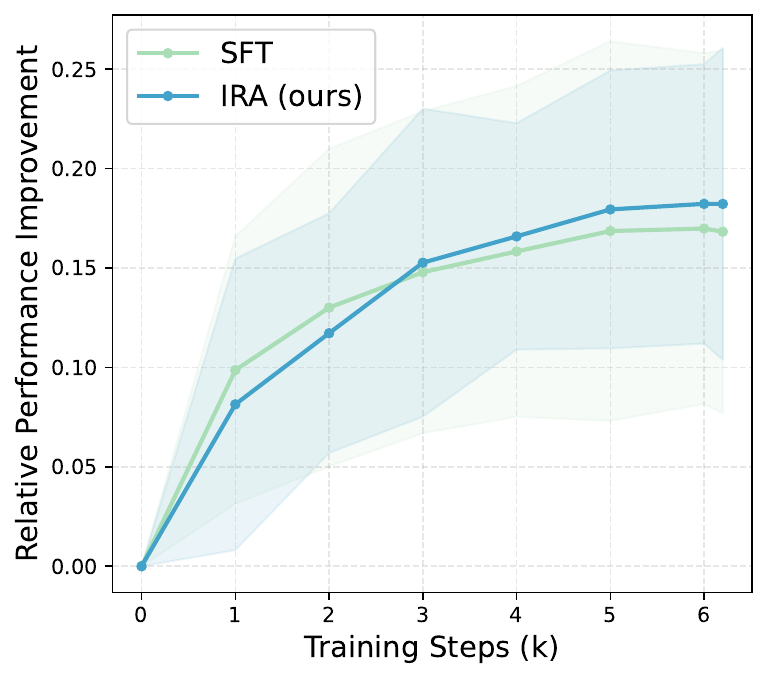}
    \caption{InternVL2-VL-8B}
  \end{subfigure}
  \hfill
  \begin{subfigure}{0.45\linewidth}
    \includegraphics[width=0.9\linewidth]{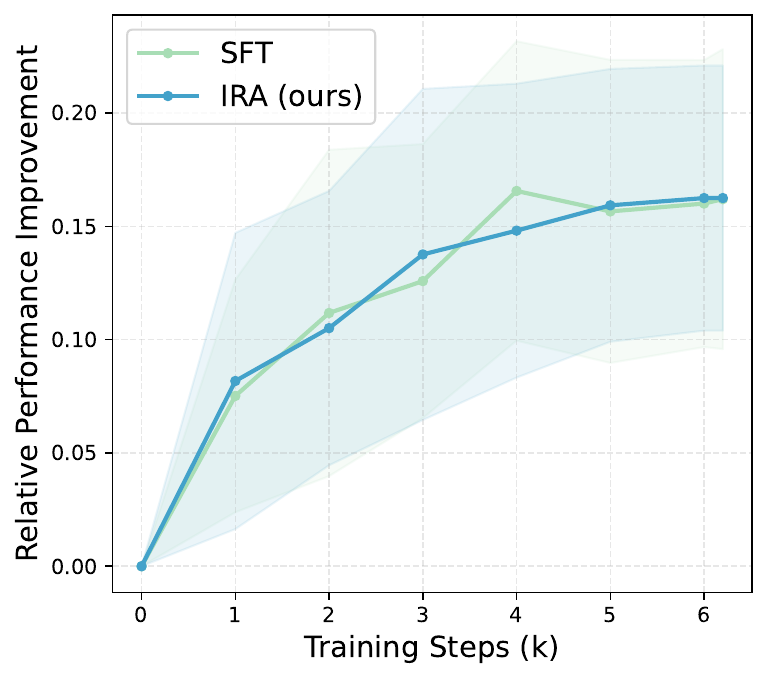}
    \caption{InternVL2.5-VL-8B}
  \end{subfigure}
  \caption{\label{fig:p_trend} Comparison of relative improvement over the baseline on 10 benchmarks during training. Overall, IRA shows a more stable increase in performance.}
\end{figure}

Supervised fine-tuning (SFT) of vision-language models is known to suffer from both catastrophic forgetting~\cite{chen-etal-2020-recall,dong2021should} and overfitting, particularly when adapting pretrained models to large-scale, task-specific visual instruction data. During SFT, the model is optimized solely for next-token prediction on the finetuning distribution, without explicitly preserving the information structure learned during pretraining. As a result, previously acquired general visual reasoning capabilities are gradually overwritten, leading to degraded generalization across diverse tasks. This phenomenon is reflected in the training dynamics shown in the Fig.~\ref{fig:p_trend}. After evaluating average performance across 18 benchmarks on two InternVL models, we observed that while SFT initially improves performance, it becomes unstable over training steps, indicating overfitting to the finetuning data. In contrast, IRA demonstrates more consistent, monotonic improvement throughout training, achieving higher relative performance gains.
This smoother optimization trajectory suggests that IRA mitigates catastrophic forgetting by regularizing visual representations, preventing the model from collapsing into narrow, task-specific ones. 
Empirically, these results show that explicitly regularizing visual representation learning in VLM can improve training stability and downstream robustness of full-parameter instruction tuning.

\begin{table*}[t]\setlength{\tabcolsep}{4pt}
\centering
\caption{Study of $\beta_{max}$ with IRA in transformer layers from depth 60\%-80\% in InternVL2-8B.}\label{tab:betaexp2}
\scalebox{0.8}{
\begin{NiceTabular}{lcccccccc}
\toprule
 $\beta_{max}$& $k$& MMMU & MMStar  & VQA$^{text}$&ChartQA & EmbSpatial & MuirBench &Avg. \\ \midrule
 $8\times10^{-5}$&0.4& 45.4&59.4&73.6&79.8&62.9&36.7&59.6\\
$1\times10^{-4}$&0.5&45.7&58.0&73.8&79.9&64.9&36.8&\textbf{60.0}\\
\bottomrule
\end{NiceTabular}}
\end{table*}

\begin{table*}[t]\setlength{\tabcolsep}{4pt}
\centering
\caption{Study of $\beta_{max}$ with IRA in transformer layers from depth 60\%-80\% in InternVL2.5-8B.}\label{tab:betaexp3}
\scalebox{0.8}{
\begin{NiceTabular}{lcccccccc}
\toprule
 $\beta_{max}$&$k$& MMMU & MMStar  & VQA$^{text}$&ChartQA & EmbSpatial & MuirBench  &Avg.\\ \midrule
 $8\times10^{-5}$&0.3&47.2&59.0&75.2&81.3&64.1&34.9&60.3 \\
 $8\times10^{-5}$&0.4&45.3&58.7&75.0&81.6&65.2&36.4&60.3 \\
$1\times10^{-4}$&0.5&47.6&58.8&74.7&81.8&64.2&35.4&\textbf{60.4}\\
\bottomrule
\end{NiceTabular}}
\end{table*}

\subsection{Study of the KL Weight}\label{appen:ana2}
In practice, directly applying noise injection to a pretrained VLM can destabilize training, since the model has already learned a highly structured embedding geometry through large-scale pretraining. Introducing a strong KL constraint at the beginning of training may therefore disrupt the learned representation space. To mitigate this issue, we gradually increase the KL weight $\beta$ using a cosine warm-up schedule, where $\beta$ is interpolated from 0 to a maximum value $\beta_{max}$ during $k\%$ of the total training steps.

We first study the effect of different $\beta_{max}$ values when applying IRA to transformer layers \{17,21\} in Table~\ref{tab:betaexp}. We then study the $\beta_{max}$ in the continuous IRA setting in Table~\ref{tab:betaexp2} and Table~\ref{tab:betaexp3}. 
Empirically, we have observed a correlation between the number of IRA layers and $\beta_{max}$. Specifically, inserting more IRA layers into a pretrained VLM requires a larger $\beta_{max}$ with more warm-up steps. 

\subsection{Limitation} 
Due to resource constraints, we apply the proposed methods to models up to 8B parameters, but we expect the conclusions to hold for larger models with more parameters, such as 13B, 26B, and 73B. Additionally, we believe that an IRA can serve as a general architecture for robust representation learning during the pre-training stage of LLMs and VLMs. We leave this to future study.

\begin{table*}[t]\setlength{\tabcolsep}{4pt}
\centering
\caption{Study of $\beta_{max}$ with IRA in transformer layer \{17, 21\}. }\label{tab:betaexp}
\scalebox{0.8}{
\begin{NiceTabular}{lcccccccc}
\toprule
 $\beta_{max}$&k& MMMU & MMStar  & VQA$^{text}$&ChartQA & EmbSpatial & MuirBench &Avg. \\ \midrule
 $2\times10^{-5}$&0.3&46.3&58.2&72.6&77.5&61.4&43.7&\textbf{60.0} \\
$4\times10^{-5}$&0.3&44.6&59.2&71.4&79.8&61.7&33.4&58.4\\
$6\times10^{-5}$&0.3&45.6&58.0&71.9&78.5&63.9&37.5&59.2\\
\bottomrule
\end{NiceTabular}}
\end{table*}

\begin{figure*}[t]
  \centering 
  \begin{subfigure}{0.32\linewidth}
  \centering
    \includegraphics[width=\linewidth]{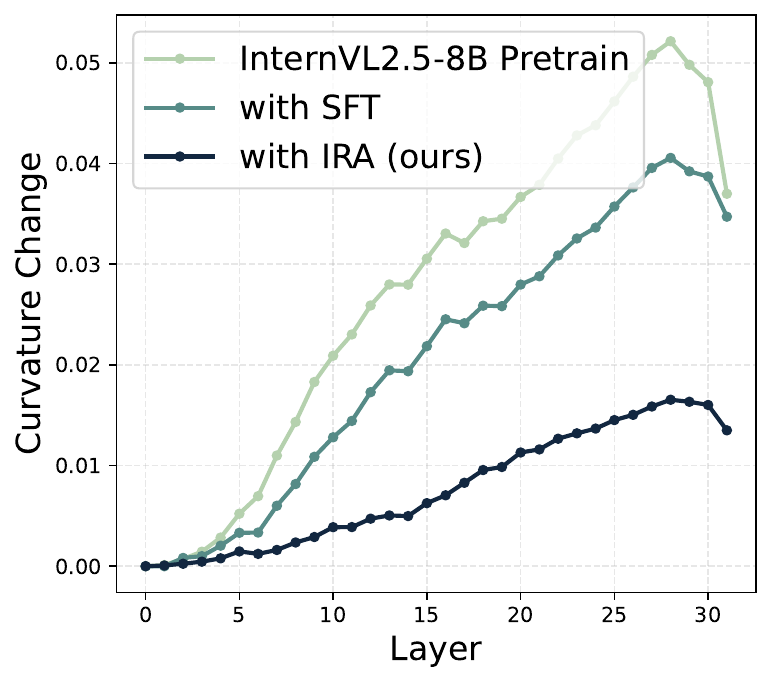}
    \caption{Curvature Change}
  \end{subfigure}
  \hfill
  \begin{subfigure}{0.32\linewidth}
  \centering
    \includegraphics[width=\linewidth]{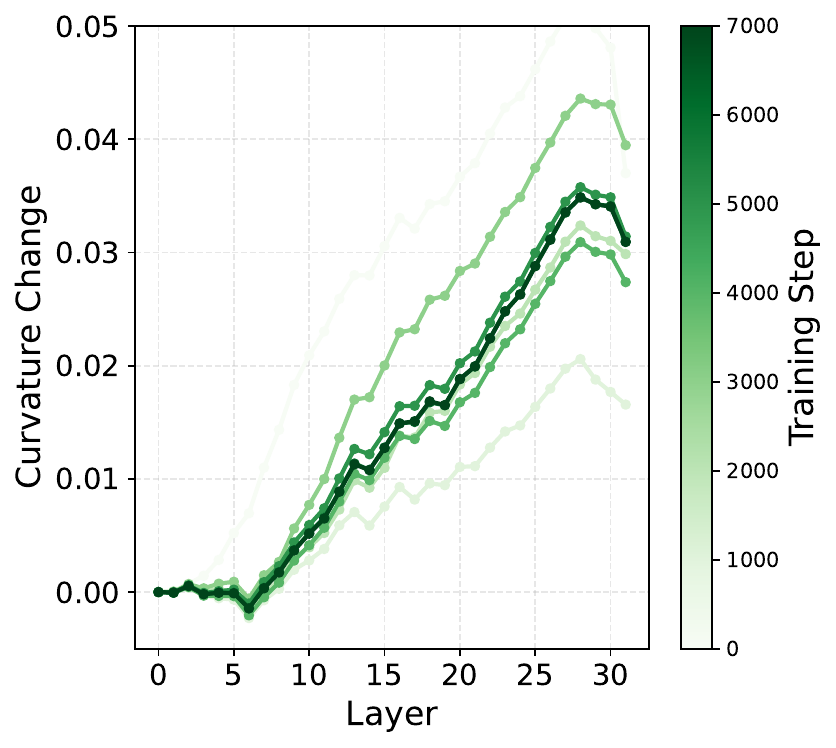}
    \caption{InternVL2-SFT}
  \end{subfigure}
  \hfill
  \begin{subfigure}{0.32\linewidth}
  \centering
    \includegraphics[width=\linewidth]{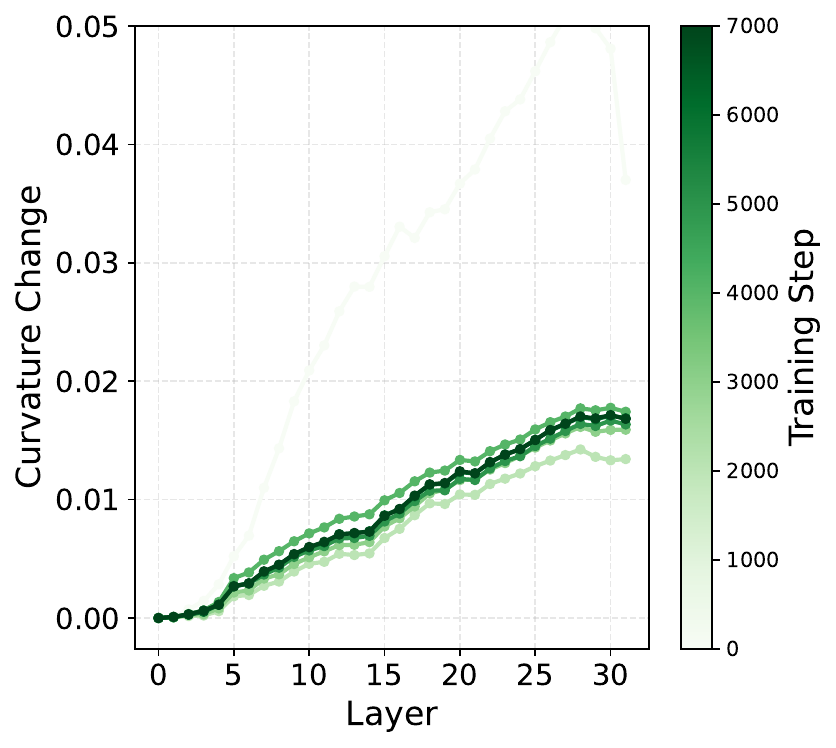}
    \caption{InternVL2-IRA}
  \end{subfigure}
  \caption{\label{fig:curvature2}Analysis of the representation straightening. The y-axis shows the $\Delta C$ of each layer. A straighter trajectory indicates a smoother update to the representation. A model with better embedding quality exhibits a straighter curvature trajectory. }
\end{figure*}

\begin{figure*}[t]
  \centering
    \includegraphics[width=0.8\linewidth]{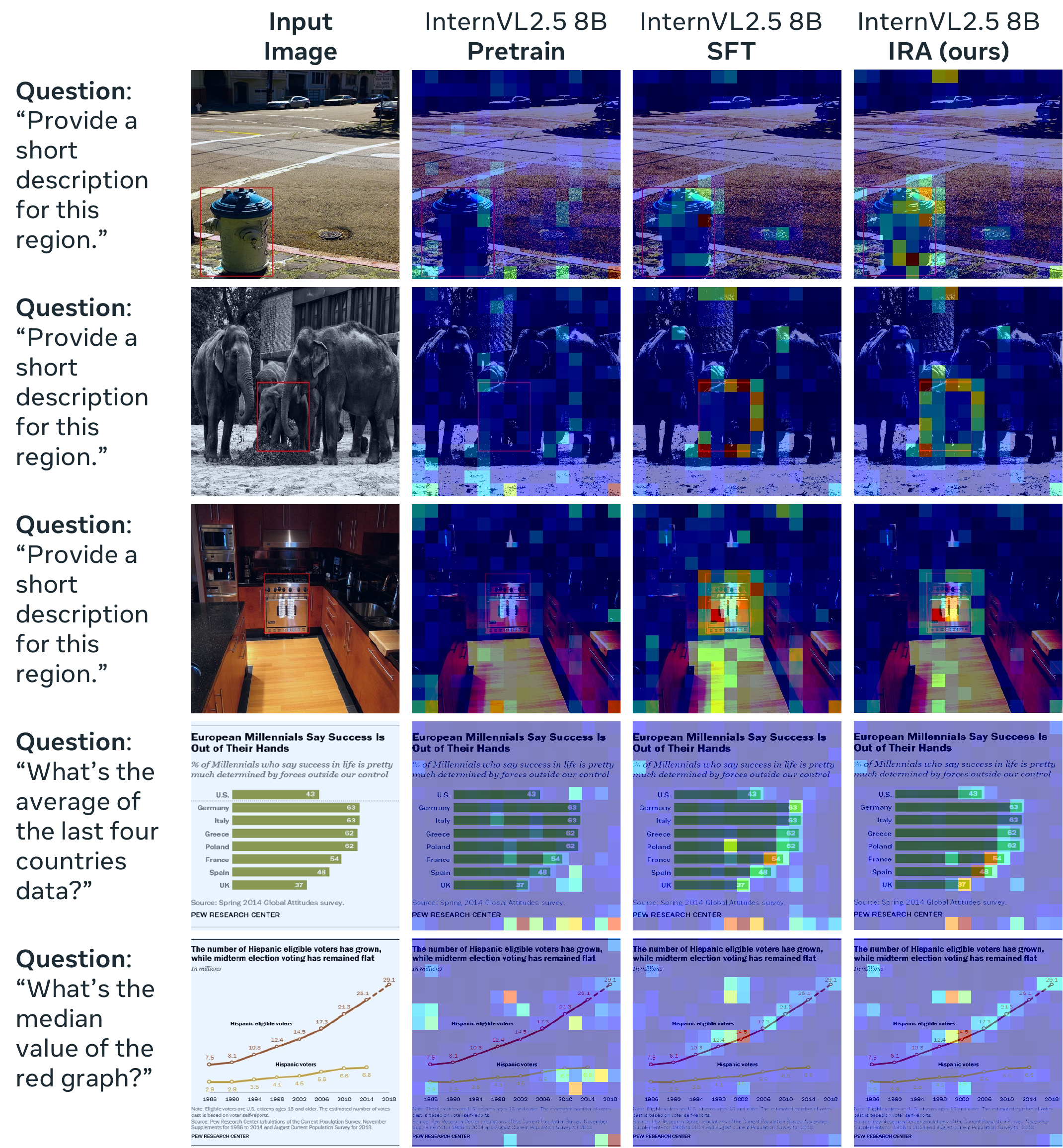}
  \caption{\label{fig:attn_vis}Visualization of token-wise attention map. The VLMs take only an image and a text instruction as input. }
\end{figure*}
\section{Experimental Results}

\subsection{Training Recipe}\label{appen:hyperp}
In this work, we use 128$\times$A100 (80G) for training and 8$\times$A100 (80G) for testing. This work primarily follows InternVL~\cite{internvl} and LLaVA-OneVision~\cite{li2024llava} in setting the hyperparameters. However, we reduce the learning rate from $4\times 10^{-5}$ to $1\times 10^{-5}$ for training the InternVL model in stage 2 due to the catastrophic forgetting issue since we observe that overly large learning rates destabilize training where the training loss of both SFT and IRA both decrease in the first few training steps and converge to a higher loss after training, leading to bad performance. 
\subsection{Analysis}\label{appen:visualization}
In Fig.~\ref{fig:attn_vis}, we provide additional visualizations of attention maps, with samples collected from RefCOCOg~\cite{refcoco+} and ChartQA~\cite{masry-etal-2022-chartqa}. After training on the same data, IRA reduces noisy visual attention and assigns higher attention weights to relevant visual tokens than SFT.

In Fig.~\ref{fig:curvature2}, we analyze the embedding quality of the InternVL2 model. We have observed the same trend when comparing with the InternVL2.5 model. This consistency suggests that IRA generalizes well across different models.

\end{document}